\documentclass[preprint, 5p, times]{elsarticle}

\usepackage{amsmath}
\usepackage{amssymb}

\usepackage{color}
\usepackage[dvipsnames]{xcolor}
\definecolor{violet_ensam}{RGB}{142,37,98}
\definecolor{orange_ensam}{RGB}{242,148,0}
\definecolor{gris_ensam}{RGB}{88,88,90}

\usepackage{blindtext}
\usepackage{graphicx}
\usepackage{multirow}
\usepackage{booktabs}
\usepackage[normal]{caption}
\usepackage{subcaption}

\usepackage{mathtools}
\usepackage{siunitx}

\usepackage{rotating}
\usepackage{adjustbox}
\usepackage{tablefootnote}
\usepackage{makecell}

\usepackage[inline]{enumitem}

\usepackage[section]{placeins}

\usepackage{tikz}
\usetikzlibrary{calc,shapes,arrows,decorations.text,positioning,patterns,matrix,chains,decorations.pathreplacing, decorations.pathmorphing}

\usetikzlibrary{external}
\tikzset{sum/.style = {draw, circle, node distance = 2cm}}

\usepackage{pgfplots}
\pgfplotsset{compat=newest}

\usepackage{threeparttable}

\usepackage[inline]{enumitem}

\makeatletter
\newcommand{\thickhline}{%
    \noalign {\ifnum 0=`}\fi \hrule height 1pt
    \futurelet \reserved@a \@xhline
}
\newcolumntype{"}{@{\hskip\tabcolsep\vrule width 1.5pt\hskip\tabcolsep}}
\makeatother

\DeclareMathOperator*{\argmax}{argmax}

\usepackage{threeparttable}

\usepackage{wrapfig}
\usepackage{scalefnt}

\usepackage[mathscr]{euscript}

\usepackage[outline]{contour}

\usepackage{relsize}

\usepackage[]{algorithm2e}

\usepackage{bookmark}

\usepackage{pdfpages}

\begin{document}
\begin{frontmatter}

\title{Combining pretrained CNN feature extractors to enhance clustering of complex natural images}

\author[label1]{Joris Gu\'erin$^*$}
\author[label2]{St\'ephane Thiery}
\author[label2]{Eric Nyiri}
\author[label2]{Olivier Gibaru}
\author[label3]{Byron Boots}

\address[label1]{Universidade Federal do Rio Grande do Norte, Natal, Brazil (corresponding: jorisguerin.research@gmail.com)}
\address[label2]{\'Ecole nationale des Arts et M\'etiers ParisTech, Lille, France}
\address[label3]{University of Washington, Seattle, USA}

\begin{abstract}
Recently, a common starting point for solving complex unsupervised image classification tasks is to use generic features, extracted with deep Convolutional Neural Networks (CNN) pretrained on a large and versatile dataset (ImageNet). However, in most research, the CNN architecture for feature extraction is chosen arbitrarily, without justification. This paper aims at providing insight on the use of pretrained CNN features for image clustering (IC). First, extensive experiments are conducted and show that, for a given dataset, the choice of the CNN architecture for feature extraction has a huge impact on the final clustering. These experiments also demonstrate that proper extractor selection for a given IC task is difficult. To solve this issue, we propose to rephrase the IC problem as a multi-view clustering (MVC) problem that considers features extracted from different architectures as different ``views'' of the same data. This approach is based on the assumption that information contained in the different CNN may be complementary, even when pretrained on the same data. We then propose a multi-input neural network architecture that is trained end-to-end to solve the MVC problem effectively. This approach is tested on nine natural image datasets, and produces state-of-the-art results for IC.
\end{abstract}

\end{frontmatter}


\section{Introduction}\label{sec:intro}


This paper addresses the problem of Image Clustering (IC), also called image-set clustering\footnote{This problem should not be confused with image segmentation \cite{segmentation}, which is also sometimes called image clustering.}, which has received a lot of attention over the last three decades \cite{webscale1, GMM, alternating_opt_clust}. It has applications for searching large image databases \cite{webscale1, webscale2, webscale3}, concept discovery in images \cite{conceptdiscovery}, storyline reconstruction \cite{storyline}, medical images classification \cite{alternating_opt_clust} and robotic sorting \cite{ijaia18}, among others. In recent research, the family of deep clustering methods~\cite{survey_e2e} has shown excellent results on datasets containing small images (MNIST, USPS). However, to obtain good clustering results on more complex IC problems, i.e. large images representing real world scenes or objects, it is usually necessary to extract features from pretrained Convolutional Neural Networks (CNN) as a preprocessing step~\cite{aifu18, webscale2, imsat}. Nevertheless, there exists a variety of publicly available pretrained CNN architectures and, to the best of our knowledge, choosing which one to use has not been studied yet. Indeed, in all the literature mentioned in Section~\ref{sec:related_work}, the choice of the pretrained CNN architecture for feature extraction is never the same, and never justified. There might be several explanations for such lack of research in this direction. First, as IC is unsupervised, it is not possible to cross-validate design choices for a specific problem, thus making the architecture selection process very challenging. Moreover, using any CNN feature extractor usually leads to a huge boost in performance compared to standard computer vision features. These excellent results might hide the fact that a good architecture choice might improve clustering performance even more. Therefore, the two main objectives of this work are to study in detail the use of deep pretrained CNN features for unsupervised classification tasks and to propose a solution to the challenging problem of feature extractor selection.

\subsection{Contributions and paper organization}

In Section~\ref{sec:related_work}, we present the relevant recent work in the field of IC. Then, the organization of this paper is two-fold: 

First, in Section~\ref{sec:benchmark}, a preliminary study, consisting of an extensive set of experiments, is conducted over various IC datasets to investigate the interrelations between the type of dataset, the CNN architecture, the feature extraction layer and the type of clustering algorithm. The results from these experiments reveal that the last layer before softmax is the best for feature extraction, for every architecture and dataset. The rest of our conclusions can be summarized in one simple sentence: properly choosing the CNN architecture is important to obtain good clustering results but it is a difficult task.

Second, relying on the results from Section~\ref{sec:benchmark}, we aim to remove the need for CNN architecture selection, which is a crucial yet challenging design choice. Following the intuition that different pretrained deep networks may contain complementary information (Section~\ref{sec:ic2mvc_intuition}), we propose to use multiple pretrained networks to generate multiple feature representations (Section~\ref{sec:ic2mvc_formulation}). Such representations are treated as different ``views'' of the data, thus casting the initial IC problem into Multi-View Clustering (MVC). The relevance of this approach is demonstrated experimentally in Section~\ref{sec:ic2mvc_expe}. Then, building on the recent successes of end-to-end clustering~\cite{survey_e2e}, we also propose to leverage JULE~\cite{jule}, a deep clustering method, to solve the MVC problem (Section~\ref{sec:dmvc}). By adapting JULE to optimize the weights of a parallel neural network architecture we demonstrate state-of-the-art IC results on several public image datasets (Section~\ref{sec:exp_mvc}). This approach to MVC also has the advantage of producing a unified representation of the initial dataset which is low-dimensional and compact. An overview of the proposed method to solve IC can be seen in Section~\ref{sec:ic2mvc} (Figure~\ref{fig:dmvc}).

The idea of using multiple pretrained CNN for feature extraction was already presented in our conference paper \cite{bmvc18}. This paper extends our previous work by establishing experimentally the importance of feature extractor selection, conveying further insights on the use of multiple CNN architectures through new experiments, and improving the experimental validation on new datasets and subproblems.

\section{Related work}\label{sec:related_work}


\subsection{Definition of Image Clustering (IC)}\label{sec:relatedWork_defIC}
\begin{figure}
\centering

\begin{subfigure}[]{0.48\textwidth}
\centering
\includegraphics[width=\textwidth]{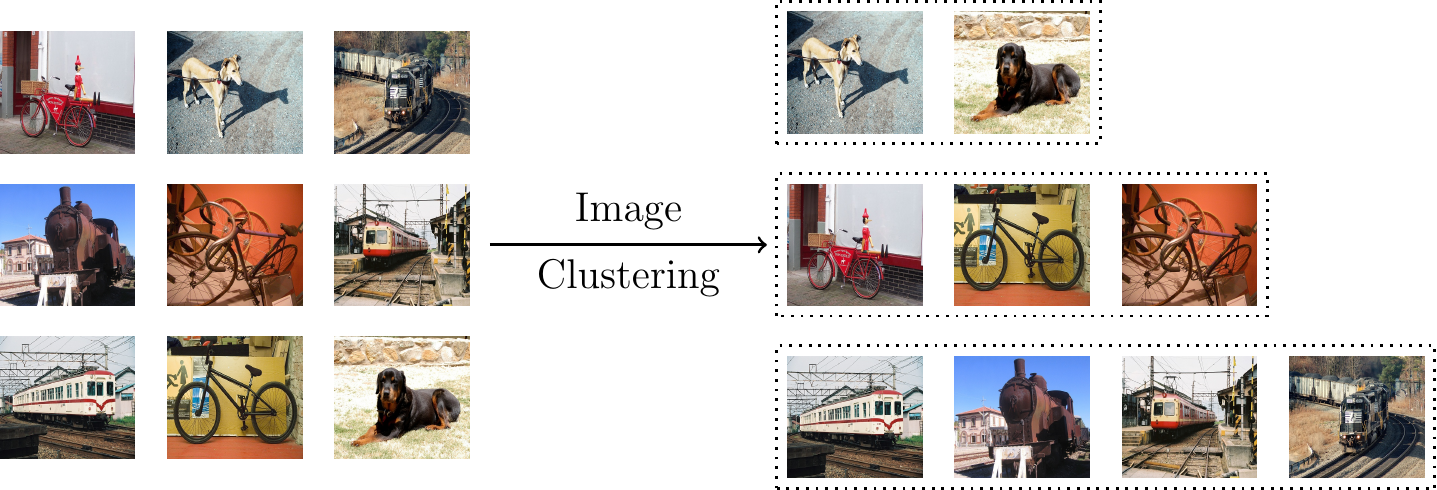}
\caption{VOC2007~\cite{voc2007}}
\label{fig:exampleIC_voc2007}
\end{subfigure}

\begin{subfigure}[]{0.48\textwidth}
\centering
\includegraphics[width=\textwidth]{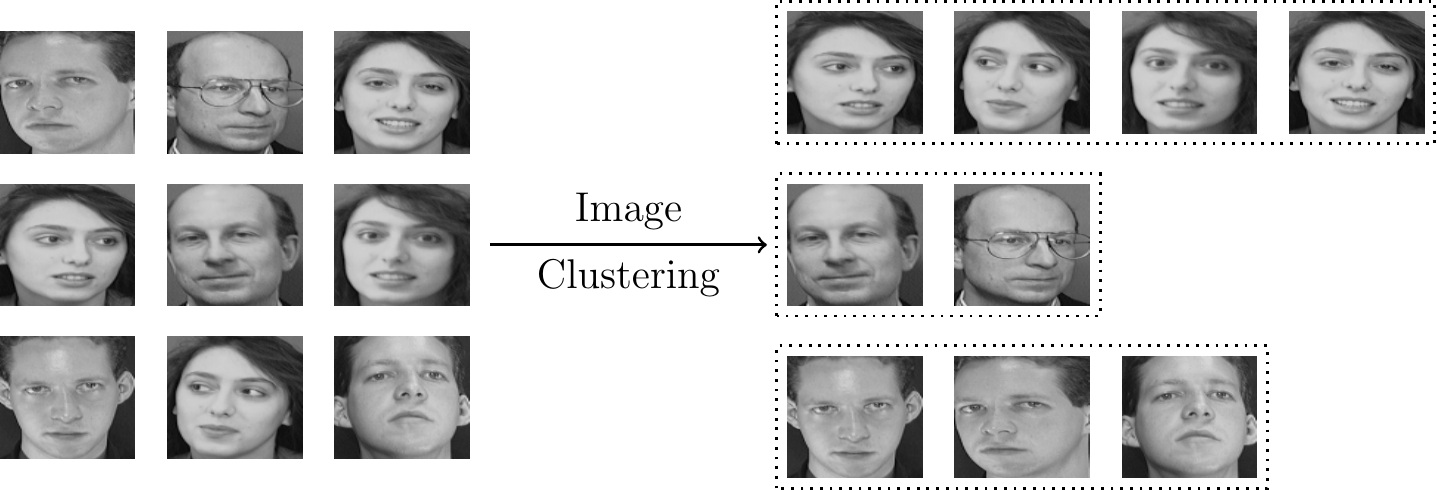}
\caption{ORL~\cite{orl}}
\label{fig:exampleIC_orl}
\end{subfigure}

\caption{Definition of the image clustering problem. Examples of inputs and expected outputs on two natural images datasets.}
\label{fig:exampleIC}
\end{figure}

Given a set of unlabeled images, the IC problem consists in finding subsets of images based on their content: two images representing the similar objects should be clustered together and separated from images representing objects of different nature. The similarity of content between images can have various definitions, for instance images can be grouped based on the dominant color, the quality of the picture, the type of background, etc. This paper studies algorithms that produce ``human-level'' classification, in other words, the objective is to reproduce the classification that human subjects performing the task would obtain. As it is hard to define ``human classification'' in an unambiguous way, we only use supervised datasets in this study. Although the labels are not used during the clustering process, they can serve as proxies for human classification during evaluation. Examples of expected outputs from an IC algorithm are illustrated in Figure~\ref{fig:exampleIC}. We also note that this study focuses on the IC setting where the number of clusters is a user defined parameter.

\subsection{End to end image clustering}\label{sec:related_work_e2eic}

Over the past four years, end-to-end clustering methods based on neural networks have produced excellent results. The two pioneer methods for deep clustering were Deep Embedding for Clustering analysis (DEC)~\cite{dec} and Joint Unsupervised LEarning of deep representations and image clusters (JULE)~\cite{jule}. A complete literature review of the topic is outside the scope of this paper, for that we refer the reader to the following recent survey~\cite{survey_e2e}. However, it is worth mentioning some recent work that obtained very strong results. The SA-Net framework~\cite{sa_net} proposes to extend spectral clustering into a deep learning framework. Multi-Modal Deep Clustering (MMDC)~\cite{mmdc} trains a neural network to align its image embedding with target points sampled from a Gaussian Mixture Model distribution. In~\cite{ddccsc}, the authors propose to rewrite the k-means clustering algorithm in a way that it can be optimized with respect to both the embedding parameters and the cluster parameters via stochastic gradient descent. The approach proposed in~\cite{ddbic} builds on density based clustering methods to propose an end-to-end method that does not need the number of clusters. In~\cite{ssdec}, pairwise constraints are incorporated to DEC in order to improve its results.

\subsection{Features used for image clustering}\label{sec:related_work_features}

The first successful methods to address IC focused on feature selection and used sophisticated algorithms to deal with complex representations. For instance, in~\cite{GMM}, images are represented by Gaussian Mixture Models fitted to the pixels and clusters the set of images using the Information Bottleneck (IB) method~\cite{info_bottleneck}. In~\cite{segmentation_image_set}, the authors use features resulting from image joint segmentation and sequential IB for clustering. The approach proposed in~\cite{commonality_clustering} consists in using bags of features with local representations (SIFT, SURF, etc.) and defining commonality measures used for agglomerative clustering. 

Recently, several deep IC methods (presented above) have demonstrated very strong unsupervised classification results on raw image data for various datasets containing small images. For example, for both MNIST (images size: 28x28) and USPS (16x16), the clustering accuracy reported in~\cite{ddccsc} are above~95\%. However, these end-to-end clustering methods tend to struggle when trying to cluster datasets of large images representing real-world objects and scenes and better results can be obtained by applying simple clustering algorithms on top of pretrained CNN features. For example, for COIL100 (128x128), which is a relatively easy dataset with high intra-cluster similarity, the state-of-the-art clustering accuracy is around~77\%~\cite{li2018discriminatively}. This difficulty to handle complex datasets of large images can also be seen in the fact that there is currently no deep clustering method addressing cases with images larger than COIL100. 

A solution to deal with this issue is to replace raw image data by features extracted from Convolutional Neural Networks (CNN) pretrained on ImageNet~\cite{imagenet}. For example, by applying a simple K-Means on top of features extracted by a pretrained ResNet50, we can obtain a clustering accuracy above~81\%, which is already above state-of-the-art with raw data. This approach has been used widely in the IC literature recently, and~\cite{infinite_ensemble, alternating_opt_clust, webscale2, aifu18, imsat} all use pretrained CNN feature extractors to generate a new data representation of the images before clustering. The adoption of pretrained CNN features has lead to considerable improvements in clustering results, however, many pretrained CNN are available online. In this paper, we conduct extensive experiments in order to gain insight about the feature extractor selection problem, which has never been studied thus far. 

\subsection{Multi-view clustering and ensemble clustering}
\label{sec:related_work_mvc_ec}

Ensemble clustering (EC) consists in combining different clustering results in order to obtain a unified, final partition of the original data with improved clustering accuracy~\cite{ensemble_survey}. It is composed of two steps: generation, which deals with the creation of a set of partitions, and consensus, where all the partitions are integrated into a better set of clusters. In contrast to EC, Multi-View Clustering (MVC), is concerned with finding a unified partition from multi-view data~\cite{survey_mvc}, which can be obtained by various sensors or represented with different descriptors. Recently, MVC has received a lot of attention. The family of methods based on Multiple Kernel K-Means~\cite{mkkm, rmkkm_used, rmkkm1} and on Multi-View Spectral Clustering~\cite{mvsc} represent strong baseline approaches to tackle MVC. In~\cite{multiview_concat}, the authors propose different loss functions applied on the concatenated views. In~\cite{multiview_latent1} and~\cite{multiview_latent2} lower dimensional subspaces are learned before clustering with standard methods. The authors of~\cite{mlsslmvc} combine Hierarchical Self-Representative Layers, and Backward Encoding Networks to improve MVC. Finally, a deep matrix factorization solution is proposed in~\cite{mvmcdm} to solve the multi-view multi-clustering problem. 

MVC and EC are closely related and have already been combined in previous work. In~\cite{multiview_ensemble}, good MVC results are attained by embedding MVC within the EC framework. The authors leveraged the different views to generate independent partitions and then used a co-association-based method to obtain the consensus. In both~\cite{generated_mv_images} and~\cite{generated_mv_genes}, generation mechanisms borrowed from EC are used to generate artificial multiple views of the data. In this paper, we propose to use multiple pretrained CNNs to generate different feature representations of an image dataset. Hence, we generate a MVC problem from an ensemble of pretrained CNN feature extractors.

\section{Pretrained CNN features for IC: preliminary benchmark study}
\label{sec:benchmark}

This section aims at answering several questions about the use of deep features for image clustering. We want to know if different CNN architectures, although pretrained on the same dataset (ImageNet), behave differently when presented a new unsupervised dataset. Another objective is to study if features should rather be extracted from the early or late layers of the CNN. The ideal scenario would be to come up with generic rules such as: when facing a particular type of dataset DS, and to optimize a given metric M, one should choose the architecture NN, extract features from layer L and cluster the new feature set with algorithm C.

\subsection{Experiment design}
\label{sec:benchmark_exp_design}

\begin{figure}
\centering

\includegraphics[width=0.48\textwidth]{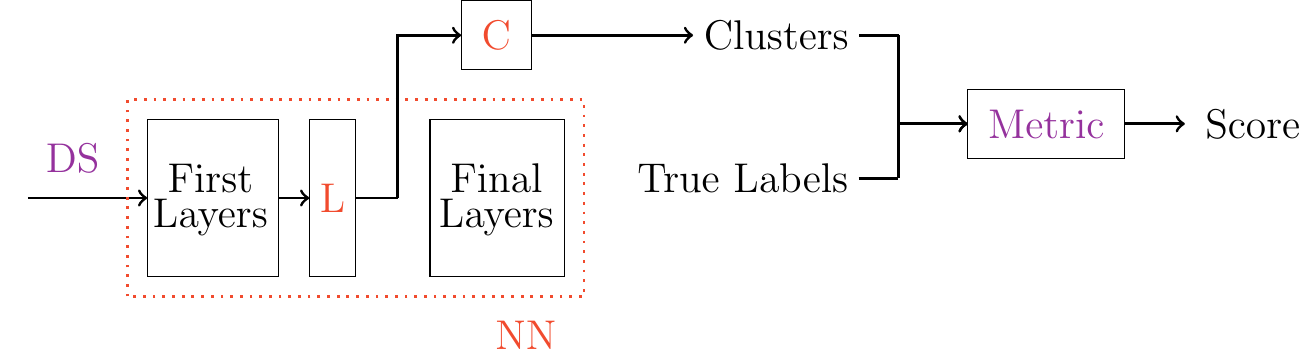}
    
\caption[General form of the proposed Image Clustering pipeline]{General form of the proposed Image Clustering pipeline.}
\label{fig:pipeline}
\end{figure}

To answer these questions, the pipeline presented in Figure~\ref{fig:pipeline} is implemented for several datasets. For each dataset we try multiple combinations of NN-L-C triplets. The results of each combination is evaluated using the Normalized Mutual Information metric. The choices made for studying the different elements in the pipeline in Figure~\ref{fig:pipeline} are described in details in the coming sections. 

\subsubsection{Datasets}
\label{sec:benchmark_exp_design_dataset}
To obtain results about CNN feature extractors which can be generalized, experiments need to be carried out on many datasets, belonging to different subtasks of IC. Hence, the proposed pipeline, with different feature extractors and clustering algorithms, is applied to the unsupervised versions of the following tasks:
\begin{itemize}
\item \textbf{Natural object recognition}: We call natural object recognition the task of classifying images based on a single object it contains. Classes are defined in the most generic way possible (cat, dog, car, etc.). This task is the most similar to ImageNet, hence, although the precise task (categories)  and domain (backgrounds) are different, pretrained deep features are expected to generate good clustering results on this task. Moreover, it is also expected that final layers will be better suited for this task as the objects to separate are similar to ImageNet classification.
\item \textbf{Scene recognition}: This task is different from what the pretrained network was trained to do. Indeed, in scene recognition, a category is defined by the simultaneous presence of multiple objects on the image. For example, a dining room needs to contain chairs and a table. We still expect the last layers to perform better feature extraction at this task as they are supposed to contain higher level information.
\item \textbf{Fine-grained recognition}: This task is also very challenging for pretrained deep features because fine-grained classes are defined within what usually defines a single category for ImageNet, and the pretrained network might have learned to produce features which are too generic for this task without additional supervision. For example, a fine-grained recognition task might consist in recognizing different species of birds.
\item \textbf{Face recognition}: This task is also a fine-grained recognition task. However, it is of such importance today that we study it as its own class of problems.
\end{itemize}

For each unsupervised classification task listed above, we pick two datasets. The datasets studied, together with their statistics, are listed in Table~\ref{tab:data_stats}. More details on these datasets can be found on the papers in which they were introduced: VOC2007~\cite{voc2007}, COIL100~\cite{coil100}, Archi~\cite{archi25}, MIT~\cite{mit67}, Flowers~\cite{flowers}, Birds~\cite{birds}, UMist~\cite{umist}, FEI~\cite{fei}.

\begin{table*}[!ht]
\caption[Datasets used for the image clustering benchmark study]{Statistics of the eight datasets used for the image clustering benchmark study.} 
\label{tab:data_stats}
\centering
\begin{threeparttable}
\scalebox{1}{
\begin{tabular}{>{\centering\arraybackslash}p{0.15\textwidth}|>{\centering\arraybackslash}p{0.15\textwidth}>{\centering\arraybackslash}p{0.1\textwidth}>{\centering\arraybackslash}p{0.1\textwidth}>{\centering\arraybackslash}p{0.15\textwidth}>{\centering\arraybackslash}p{0.1\textwidth}}

Dataset & Task & \# images & \# classes & Images size & Balanced\tnote{1} \tabularnewline \hline
VOC2007\tnote{2}
& Natural object & 2841 & 20 & Variable & No \tabularnewline
COIL100 & Natural object & 7200 & 100 & 128 $\times$ 128 & Yes \tabularnewline
Archi & Scene & 4794 & 25 & variable & No \tabularnewline
MIT & Scene & 15620 & 67 & variable & No \tabularnewline
Flowers & Fine-grained & 400 & 17 & variable & Yes \tabularnewline
Birds & Fine grained & 2800 & 200 & variable & No \tabularnewline
UMist & Face & 564 & 20 & 220 $\times$ 220 & Yes \tabularnewline
FEI & Face & 6033 & 200 & 640 $\times$ 480 & Yes

\end{tabular}}

\begin{tablenotes}
\item[1] \small \textit{A dataset is balanced if it contains a similar number of instances for each classes.}
\item[2] \small \textit{We use a modified version of the VOC2007 test set. All the images presenting two or more labels have been removed in order to be able to evaluate clustering.}
\end{tablenotes}
\end{threeparttable}
\end{table*}

\subsubsection{Architectures}
\label{sec:benchmark_exp_design_archi}

To ease and speed up development, we compare the Keras~\cite{keras} implementations of five popular CNN architectures:
\begin{itemize}
    \item Two VGG architectures: VGG16 \& VGG19~\cite{vgg},
    \item One ResNet architecture: ResNet50~\cite{resnet},
    \item Two Inception-like architectures: InceptionV3~\cite{inception}, Xception~\cite{xception}.
\end{itemize}

We also use the ImageNet pretrained weights provided by Keras. As of today, ImageNet is the only very large labelled public dataset which has enough variety in its classes to be a good feature extractor for a variety of tasks. Moreover, there are plenty of CNNs pretrained on ImageNet already available online. However, the results presented are expected to apply to CNNs pretrained on other databases, when larger and more diverse datasets will be created.

\subsubsection{Layers}
\label{sec:benchmark_exp_design_layer}

The IC problem studied in this paper consists in discovering classes represented by ``objects'' present on the pictures. Thus, the feature extractor needs to gather semantic level information about the data to make such clustering possible. Such high level information is present in the final layers of the pretrained networks. Thus, to study the impact of the layer chosen, we pick three layers among the last ones for each network : 
\begin{itemize}
    \item One layer in the end of the convolutional block (L1),
    \item The second layer before the ImageNet softmax layer (L2),
    \item The last layer before softmax (L3).
\end{itemize}
Picking layers from earlier stages of the network is both not very relevant and not practical. Indeed, the closer to the beginning of the network the layer is, the bigger the feature space is and the longer the clustering is. It is probably not relevant because the features are too low level to be informative without additional supervision.

On the one hand, the last one or two layers might provide better results as their goal is to separate the data (at least for the fully-connected layers). On the other hand, the opposite intuition is also relevant as we can imagine that these layers are too specialized to be transferable. The names (as given in the Keras implementation) of the layers retained for this study, as well as the size of their output space, are reported in Table~\ref{tab:svtc_layers}.

\begin{table*}[!ht]
\caption[Names of the feature extraction layers studied]{Names of the feature extraction layers studied.} 
\label{tab:svtc_layers}
\centering

\begin{adjustbox}{center}
\begin{tabular}{>{\centering\arraybackslash}p{0.05\textwidth}>{\centering\arraybackslash}p{0.05\textwidth}|>{\centering\arraybackslash}p{0.13\textwidth}>{\centering\arraybackslash}p{0.13\textwidth}>{\centering\arraybackslash}p{0.13\textwidth}>{\centering\arraybackslash}p{0.15\textwidth}>{\centering\arraybackslash}p{0.15\textwidth}}

& & VGG16 & VGG19 & Inception & Xception & Resnet50 \tabularnewline

\hline

\multirow{2}{*}{L1} & name & {\small block5\_pool} & {\small block5\_pool} & {\small mixed7} & {\small add\_12} & {\small activation\_40} \tabularnewline
& shape & 25,088 & 25,088 & 221,952 & 102,400 & 200,704 \tabularnewline

\hline

\multirow{2}{*}{L2} & name & {\small fc1} & {\small fc1} & {\small mixed10} & {\small block14\_sepconv2\_act} & {\small activation\_47} \tabularnewline
& shape & 4,096 & 4,096 & 131,072 & 204,800 & 25,088 \tabularnewline

\hline

\multirow{2}{*}{L3} & name & {\small fc2} & {\small fc2} & {\small avg\_pool} & {\small avg\_pool} & {\small avg\_pool} \tabularnewline
& shape & 4,096 & 4,096 & 2,048 & 2,048 & 2,048 \tabularnewline

\end{tabular}
\end{adjustbox}

\end{table*}

\subsubsection{Clustering algorithms}
\label{sec:benchmark_exp_design_algo}

Over the last fifty years, many clustering algorithms have been developed. Different surveys propose different classifications of the clustering methods~\cite{clustering_survey1, clustering_survey2}. However, a common bipartite classification of the different algorithms seem to emerge. The first group of algorithms are called partitioning methods. Data points are considered independently, as points in the feature space, and the clusters are created by separating the space into different areas. The other type of algorithms are called graph-based methods (or connectivity based methods) and consist in viewing the data as nodes on a graph, connected by a certain distance. 

In this preliminary study, the goal is not to find a good algorithm to solve a specific problem but to study the influence of the chosen CNN feature extractor (architecture + layer) on the IC results. Hence, we only consider standard clustering algorithms in order to isolate the influence of the features. Recent deep clustering methods, will be considered in Section~\ref{sec:dmvc}. To keep our experiments simple and understandable, we pick the most widely used algorithms from each of these two main families of algorithms:
\begin{itemize}
    \item K-means (KM)~\cite{kmeans_pp},
    \item Agglomerative Hierarchical Clustering (Agg)~\cite{agglomerative}.
\end{itemize}
For both algorithms, we use the default configuration of the scikit-learn implementations~\cite{sklearn}. This avoids specific fine-tuning of the clustering algorithms for deep pretrained features. 
There exists a variety of other simple and very popular clustering methods~\cite{clustering_survey1, clustering_survey2}. The ones available in scikit-learn have been tried~\cite{aifu18} and did not present a major interest with respect to the conclusions drawn from these experiments. However, keeping connectivity based and graph based algorithms enables us to analyze if different architectures represent data differently.

\subsubsection{Metrics}
\label{sec:benchmark_exp_design_metric}

Although labels are not used for clustering, the proposed experiments are carried out on datasets that are inherently supervised. Hence, an external validation metric~\cite{external_clustering_val} can be used to evaluate the quality of the clustering for the different combinations. For this preliminary study, we choose to analyze the clustering results using Normalized Mutual Information (NMI), which is an information theoretic based metric, defined as:
\begin{equation}
    NMI(Y,C) = \frac{2 \times I(Y,C)}{H(Y) + H(C)},
\end{equation}
where $Y$ is the list of ground truth labels, $C$ the cluster assignments, $H(.)$ represents the entropy and $I(Y, C)$ the mutual information between $Y$ and $C$. NMI ranges between $0$ and $1$, with $1$ representing perfect accuracy. 

NMI is very commonly used in the literature about clustering and seem to be a good option for this preliminary study. In the second part of this paper, to study the full proposed pipeline, we also use the Cluster purity (PUR) and Clustering Accuracy (ACC) metrics to validate the results obtained. For convenience, these metrics are introduced here:
\begin{equation}
    PUR(Y,C) = \frac{1}{N} \sum_{c \in C} \max_{y \in Y} \mid c \cap y \mid,
\end{equation}
were $N$ is the number of elements in the dataset. Purity measures how much each cluster contains a single class, it also varies between $0$ and $1$, and a good algorithm has a purity close to $1$.
\begin{equation}
    ACC(Y,C)=\max_{perm \in P} \frac{1}{N} \sum_{i=0}^{n-1} 1\left(perm\left(C_i\right)=Y_i\right),
\end{equation}
where $1(.)$ is the indicator function and $P$ is the set of all permutations in $\left[1;K\right]$ where $K$ is the number of clusters. Clustering Accuracy finds the best cluster-classes match and use it to compute the standard accuracy. Similarly, it varies between $0$~and~$1$.

\subsection{Results}
\label{sec:benchmark_results}

All the possible combinations of the different elements composing the pipeline presented in Figure~\ref{fig:pipeline} are tested. Due to the high number of experiments carried out in this benchmark study, the complete results are only presented in the Appendix to improve clarity. The full tables of results for our experiments can be found in~\ref{annex:full_results_svtc}, they contain the NMI scores and clustering time for the eight datasets. 
The body of this section only presents a summary of these results in order to highlight the important information.

For completeness, \ref{annex:full_results_svtc} also includes results using bag of sifts features (BoF) representations~\cite{commonality_clustering}. This enables to compare CNN features with standard computer vision features and we can see that the NMI scores are much below deep features. We note that BoF results only appear for the smallest dataset of each task because BoFs are expensive to compute and of limited interest for this study.

To evaluate the influence of specific components of the clustering pipeline, we consider our experiments as a 4-dimensional design space which dimensions are: architecture, layer, task, clustering algorithm. The correlation between two factors is then studied by computing the mean and standard deviation (std) over all the results containing them. These two statistics enable us to evaluate both the overall performance and the stability of a combination. Although our experiments are relatively small to draw general conclusion, they enable to give a general trend.

\subsubsection{Influence of the layer}
\label{sec:benchmark_results_layer}

\begin{figure}[!ht]
\captionsetup[subfigure]{justification=centering}
\centering

\begin{subfigure}[b]{0.48\textwidth}
\centering
\includegraphics[width=\textwidth]{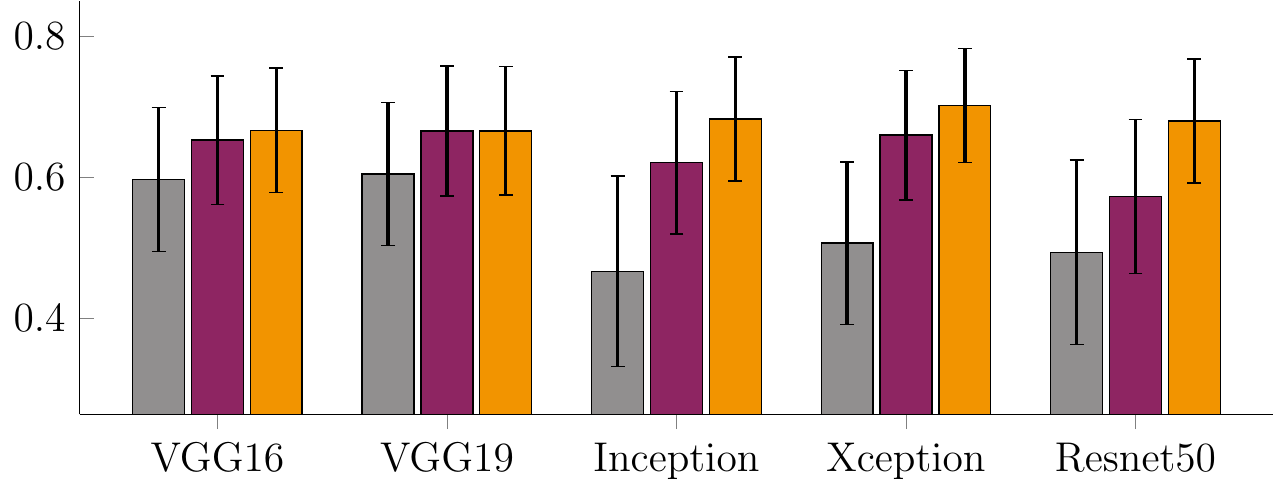}

\caption{Layer-architecture interaction \\ (mean and std of NMI across tasks and clustering algorithms).}
\label{fig:svtc_results_layer_archi}
\end{subfigure}

\begin{subfigure}[b]{0.36\textwidth}
\centering
\includegraphics[width=\textwidth]{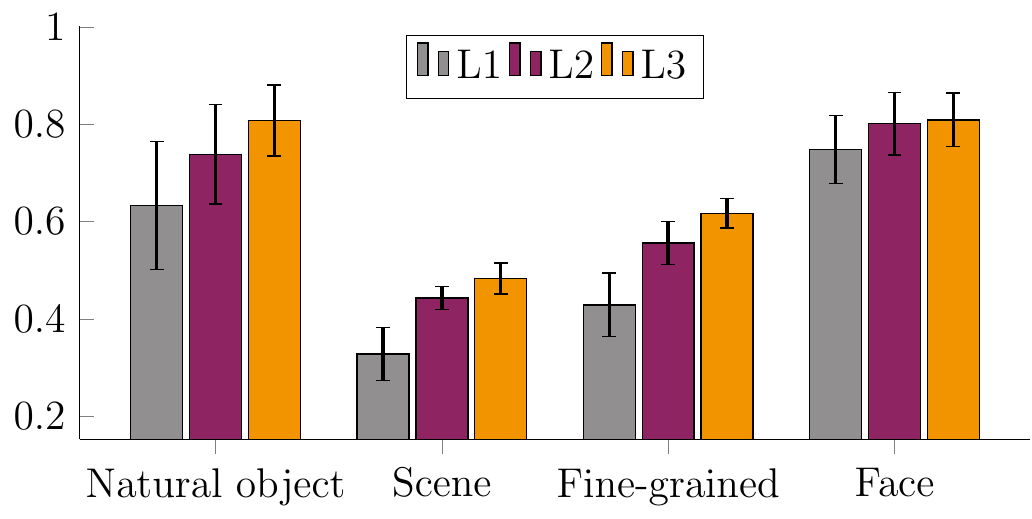}

\caption{Layer-task interaction \\ (mean and std of NMI across architectures, datasets and clustering algorithms).}
\label{fig:svtc_results_layer_task}
\end{subfigure}

\caption{Influence of the layers on the clustering results.}
\label{fig:svtc_results_layer}
\end{figure}

This results section begins by studying the impact of the choice of the layer on the clustering results. We want to know how different architectures perform under different layers. The relation between the clustering task and the position of the layer in the network is also studied. Figure~\ref{fig:svtc_results_layer} presents a summary of our experimental results regarding the impact of the layer chosen.

The results under the different architecture-layer pairs can be visualized in Figure~\ref{fig:svtc_results_layer_archi}. For all architectures, NMI scores have higher mean and lower standard deviation for later layers. This reveals that, for our experiments, final layers perform better overall and are more consistent. The high standard deviation of earlier layers shows that in some cases, features extracted from early layers can perform well, however, there is more variability and the results can drop to much lower scores in other cases. Such statement can be analyzed in more detailed looking at the results in~\ref{annex:full_results_svtc}. For example, for face recognition, some L1 layers present slightly better results than other layers, on the other hand, for fine-grained recognition, choosing L1 can results in NMI scores lower than L3 by about 0.35.

The influence of the layer on different image clustering tasks is represented in Figure~\ref{fig:svtc_results_layer_task}. Before conducting these experiments, our intuition  was that early layers would be better suited for face recognition and fine-grained classification tasks while late layers would be better at natural object and scene recognition. Indeed, high level information about an image are contained in the last layers while early layers represent lower level information (Gabor filters, color blobs, etc.)~\cite{supervised_transfer2}. However, our experimental results show that whatever the task is, later layers perform better. This effect is damped for faces but it is still true. Moreover, for all tasks, std is higher for early layers. 

These results suggest that only the last layer before softmax (L3) should be considered for clustering. Although in some cases other layers slightly outperform L3 (e.g., L2 for Xception on FEI), the profit is small and the risk is high (high std). Such finding implies that some ``low-level'' information is contained in the last layers of deep CNNs pretrained on ImageNet. This may be explained by the presence of fine-grained recognition classes within the ImageNet dataset (e.g., different breeds of dogs). Hence, only L3 layers are considered in the rest of this results section. Dropping the first layers is also motivated by the fact that their feature spaces are of higher dimensions, which means higher clustering time. For example, on average, clustering L1 layers takes about one hour while L3 layers only takes one minute. This difference is proportional to the size of the dataset.

\subsubsection{Influence of the architecture}
\label{sec:benchmark_results_archi}

\begin{figure}
\centering

\begin{subfigure}[b]{0.48\textwidth}
\centering

\includegraphics[width=0.5\textwidth]{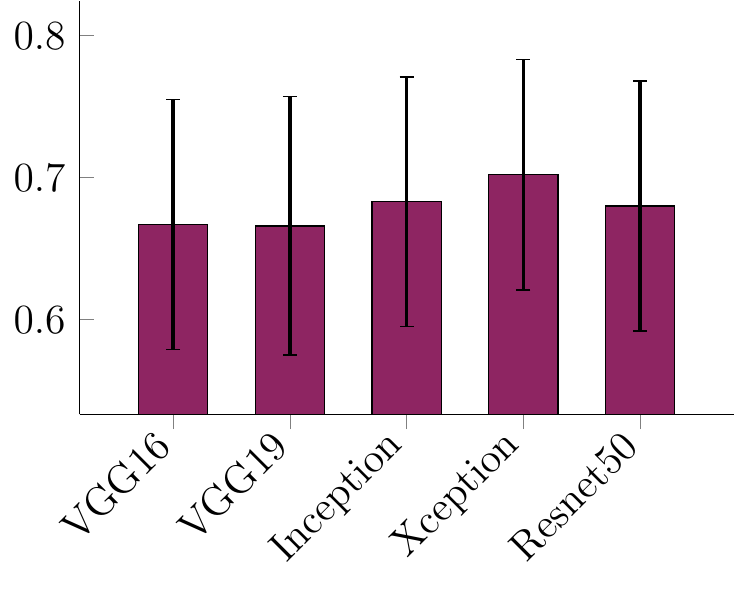}
\caption{Overall results \\ (mean and std of NMI across tasks and clustering algorithms).}
\label{fig:svtc_results_archi_general}
\end{subfigure}

\begin{subfigure}[b]{0.48\textwidth}
\centering

\includegraphics[width=\textwidth]{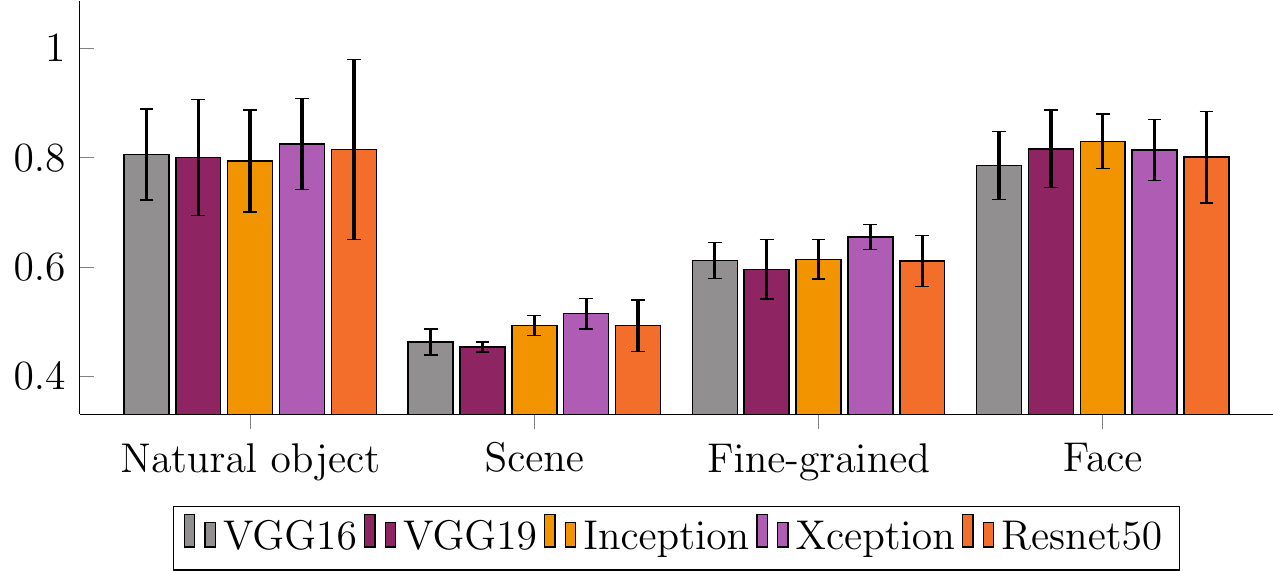}
\caption{Architecture-task interaction \\ (mean and std of NMI across datasets and clustering algorithms)}
\label{fig:svtc_results_archi_task}
\end{subfigure}

\caption{Influence of the CNN architectures (L3) on the clustering results.}
\label{fig:svtc_results_archi}
\end{figure}

The next analyses focus on the choice of the CNN architecture. These results can be found in Figure~\ref{fig:svtc_results_archi}. We want to know if an architecture is better suited for clustering than the others in general (Figure~\ref{fig:svtc_results_archi_general}) and depending on the task at hand (Figure~\ref{fig:svtc_results_archi_task}).

\begin{figure}

\centering

\begin{subfigure}[b]{0.23\textwidth}
\centering
\includegraphics[width=\textwidth]{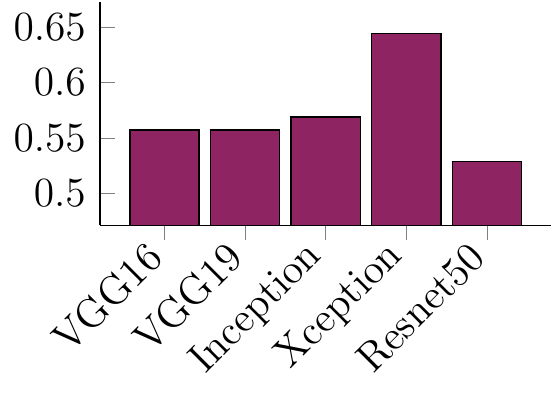}
\caption{Birds - Agg (NMI)}
\label{fig:examples_archi_birds}
\end{subfigure}
~
\begin{subfigure}[b]{0.23\textwidth}
\centering
\includegraphics[width=\textwidth]{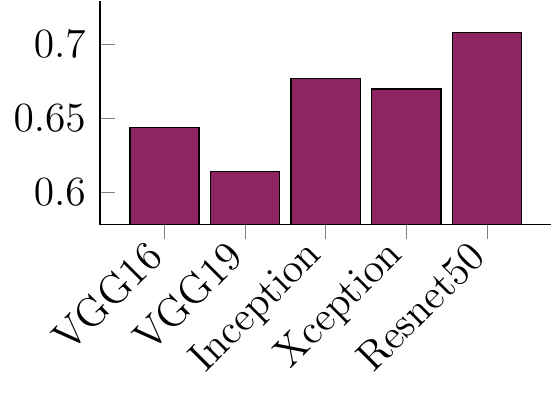}
\caption{Flowers - Agg (NMI)}
\label{fig:examples_archi_flowers}
\end{subfigure}

\caption[Criticality of the CNN architecture choice]{Examples where the choice of the CNN architecture is crucial for the clustering results.}
\label{fig:examples_archi}
\end{figure}

Besides the fact that, in our experiments, Xception presents slightly better results than the other architectures (higher mean and lower std), it is difficult to come up with relevant comments about these histograms. The results for each subtask contain too much variability, which prevents any kind of conclusion like: "for task T, use architecture A". However, we underline that an absence of strong pattern does not mean that any choice is equivalent. Indeed, there can be a huge difference in the results between a good and a bad architecture choice (Figure~\ref{fig:examples_archi}). Such absence of trend, together with the criticality of this choice, motivates the development of a new IC algorithm in Section~\ref{sec:ic2mvc}, which leverages ensemble methods to remove the need for architecture selection.

\subsubsection{Influence of the clustering algorithm}
\label{sec:benchmark_results_algo}

\begin{figure}
\centering

\includegraphics[width=0.48\textwidth]{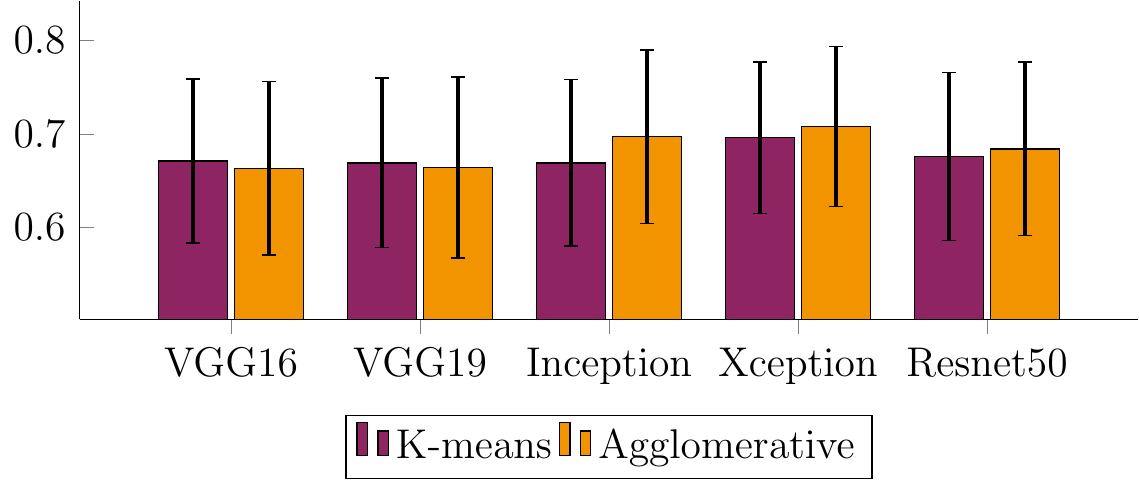}
\caption{Influence of the clustering algorithm for different CNNs (NMI).}
\label{fig:results_algos}
\end{figure}

For completeness, correlations between the architecture used and the type of clustering algorithm are also checked. This analysis is intended to investigate if the features extracted by different architecture are better suited for partitioning or graph-based methods. These results are summarized in Figure~\ref{fig:results_algos} and they do not suggest any clear conclusions. Indeed, the difference in the mean results are way smaller than the standard deviations. In other words, the risk of having poor clustering results is higher than the relative advantage one algorithm can have over the other.

\vspace{-10pt}\section{Combining CNN architectures}
\label{sec:ic2mvc}

The experiments conducted in the previous section revealed that the last layer before softmax seems to produce the most discriminative features for clustering. Our experiments also demonstrate that the selected deep feature extractor has a major impact on the results, however, they do not give any insight about how to select it. The importance of architecture selection, together with the absence of method to solve this problem, motivates the introduction of a new ensemble method to address the IC problem. In this section we propose to use different feature extractors jointly to solve IC. 

\vspace{-5pt}\subsection{Intuition}
\label{sec:ic2mvc_intuition}

\subsubsection{Intuition based on previous results}
\label{sec:ic2mvc_intuition_prevResults}

Combining CNN architectures that were pretrained on the same dataset might seem counter-intuitive as one can expect that all networks have learned the same information. This section aims at explaining the intuition behind trying this idea. This intuition is then validated experimentally in the rest of this paper.

Let $\mathcal{I} = \{0, ... 255\}^{\nu_1 \times \nu_2 \times 3}$ be the space of $\nu_1$ by $\nu_2$ colored images considered for IC. Then, a classification task T$=$(L, $f^*$) is defined by:
\begin{itemize}
    \item A set of possible labels $\text{L} = \{0, 1, ... K\}$, where $0$ represents ``none of the defined labels''.
    \item An oracle labelling function $f^*: \mathcal{I} \rightarrow \text{L}$, which associates a label to every image.
\end{itemize} 
For example, for the task of classifying images of cats and dogs, L would be $\{0, 1, 2\}$, and for an image $x$, the oracle $f^*(x)$ would output $1$ if there is a cat on the image, $2$ if there is a dog and $0$ if there is either none or both. This definition of a classification task is valid for supervised classification or unsupervised classification with known number of classes, which is the case studied in this paper. Although in practice we only study datasets composed of images which possess at least one of the labels, adding the ``zero'' label, allows us to define $\text{T}$ on all $\mathcal{I}$. 

This definition of a classification task is abstract and $f^*$ is unknown and exists independently of any dataset. In practice, to solve $\text{T}$, one first need to materialize it in the form of a dataset $\text{DS} = (X, y^*)$, where $X \subset \mathcal{I}$ is a set of images and $y^* = f^*(X)$ are the corresponding labels in L, which are inferred by human experts. For certain problems, such as medical image annotation, human experts might be scarce, thus making labelling very costly. The classification problem (T, DS) is supervised if $y^*$ is known and unsupervised else. In the supervised setting, solving $\text{T}$ for $\text{DS}$ means finding a function $f: \mathcal{I} \rightarrow \text{L}$ for which there exists a domain on which it is equal to $f^*$: $\mathscr{D}_f = \{x \in \mathcal{I} \mid f(x) = f^*(x)\}$. We call $\mathscr{D}_f$ the domain of validity of $f$. Then, we have:
\begin{itemize}
    \item $\mathscr{D}_f \subset X \Rightarrow f$ does not fit the training set,
    \item $\mathscr{D}_f = X \Rightarrow f$ overfits the dataset $\text{DS}$,
    \item $\mathscr{D}_f \supset X \Rightarrow f$ generalizes to some extent.
\end{itemize}

For many image classification problems, it is very hard to learn $f$ from scratch. Instead, it is more common to use a pretrained CNN feature extractor $f_z$ to project the initial dataset $X$ to a latent feature space of lower dimension $d$: 
\begin{equation}
\begin{array}{cccc}
f_z: & \mathcal{I} & \rightarrow & \mathbb{R}^d \\
& X & \mapsto & Z.
\end{array}
\end{equation}

\begin{figure}
    \centering
    
\begin{subfigure}[b]{0.23\textwidth}
\centering

\includegraphics[width=\textwidth]{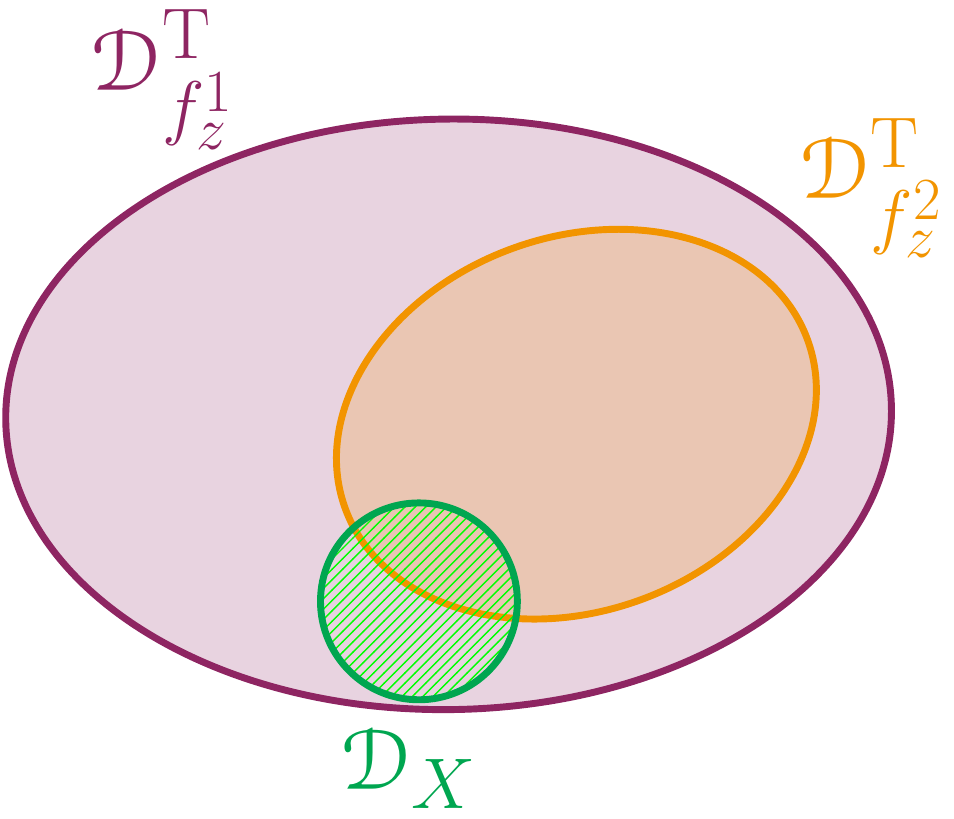}

\caption{$\mathscr{D}_{f_z^2}^{\text{T}} \subsetneq \mathscr{D}_{f_z^1}^{\text{T}}$}
\label{fig:intuitionEnsembles_inclusion}
\end{subfigure}
~
\begin{subfigure}[b]{0.23\textwidth}
\centering

\includegraphics[width=\textwidth]{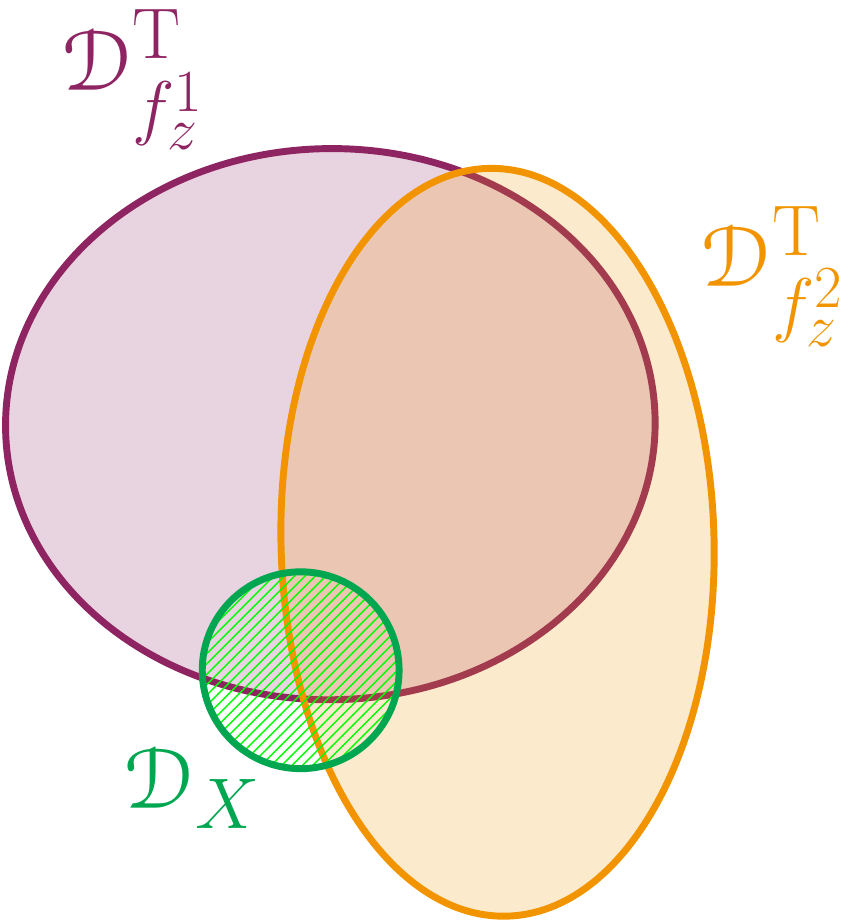}

\caption{$\forall j \in \{1, 2\}, \mathscr{D}_{f_z^j}^{\text{T}} \subsetneq (\mathscr{D}_{f_z^1}^{\text{T}} \cup \mathscr{D}_{f_z^2}^{\text{T}})$}
\label{fig:intuitionEnsembles_union}
\end{subfigure}

\caption[Intuition about multi-CNN transfer clustering]{Schematic representation of the two possible implications of $f_z^1$ being better than $f_z^2$.}
\label{fig:intuitionEnsembles}
\end{figure}

From now on, $X$ will denote a clustering dataset. Now, let $A$ be a clustering algorithm. Unlike in the supervised case, a clustering algorithm solves an unsupervised classification problem by looking at the whole set at once. In other words, if $X$ contains $N$ data points and $A$ is applied to the outputs of $f_z$, we have:
\begin{equation}
\begin{array}{cccc}
A: & (\mathbb{R}^d)^N & \rightarrow & L^N \\
& Z & \mapsto & y,
\end{array}
\end{equation}
where $y$ are the cluster assignments produced by $A$. From these definitions, we can introduce $\mathscr{D}_{f_z, A}^{\text{T}} \subset \mathcal{I}$, the domain on which applying classification algorithm $A$ to the outputs of $f_z$ enables to solve task $T$. In the previous section, our experimental results suggested that, for a given dataset, the ranking of the different feature extractors is little dependent on the chosen clustering algorithm. Hence, the subscript $A$ is dropped to define $\mathscr{D}_{f_z}^{\text{T}}$, the domain on which $f_z$ produces a ``clustering friendly'' latent space for task $\text{T}$.
Finally, let $f_z^1$ and $f_z^2$ be two pretrained CNN feature extractors, and let $\text{T}$ be the task that we aim to solve on the clustering dataset $X$. In the previous section, it was shown that $f_z^1$ and $f_z^2$ perform differently on the different datasets. For $(\text{T}, X)$, let's assume that, according to some clustering validation metric (e.g. NMI), $f_z^1$ outperforms $f_z^2$. Then, we can conclude that either
\begin{itemize}
    \item $\mathscr{D}_{f_z^2}^{\text{T}} \subsetneq \mathscr{D}_{f_z^1}^{\text{T}}$ (Figure~\ref{fig:intuitionEnsembles_inclusion}) or,
    \item $\forall j \in \{1, 2\}, \mathscr{D}_{f_z^j}^{\text{T}} \subsetneq (\mathscr{D}_{f_z^1}^{\text{T}} \cup \mathscr{D}_{f_z^2}^{\text{T}})$ (Figure~\ref{fig:intuitionEnsembles_union}),
\end{itemize}
where the $\subsetneq$ symbol represents strict inclusions.

\newpage In both situation, leveraging both networks can have positive implications on the clustering results:
\begin{itemize}
    \item \textbf{Case 1:} $f_z^1$ alone contains all the information to cluster $X$. However, in Section~\ref{sec:benchmark}, we showed that for unsupervised datasets, it is not possible to know which network performs best. Hence, using the two networks allows to make sure that all the information available for clustering are provided to the final clustering algorithm (Figure~\ref{fig:intuitionEnsembles_inclusion}).
    \item \textbf{Case 2:} The combination of the two networks contains more information than each of the networks separately. In other words, even if $f_z^2$ performs worse than $f_z^1$, it still contains information that $f_z^1$ does not. This kind of situation defines a typical setting where ensemble learning would be beneficial (Figure~\ref{fig:intuitionEnsembles_union}).
\end{itemize} 
The domain of $X$, called $\mathscr{D}_{X}$, is represented in green on both sketches of Figure~\ref{fig:intuitionEnsembles} and illustrates the two potential benefits of combining CNN feature extractors. Obviously, this intuition can be generalized to more than two networks. Although all this development is just an intuition and does not have theoretical evidences, we intend to validate it experimentally in the rest of the section.

The potential improvement from using multiple pretrained CNN feature extractors can also be understood through the following contrived example. To recognize a car, one network might learn a wheel detector while another one might detect wing mirrors. Both sets of discriminative features would enable to solve the ImageNet classification task, on which both networks were trained, but would also carry very different information that might be useful in solving a new IC task. 

Finally, it is also important to note that the opposite intuition may also be valid. Indeed, introducing redundancy of information might hide the important information and decrease the clustering results. This will be investigated in the rest of the section.

\subsubsection{Visualisation}
\label{sec:ic2mvc_intuition_visu}

To visualize this intuition on real data, we leverage the Fowlkes-Mallows Index (FMI) \cite{fm}, another external clustering validation metric, which has the advantage of having a local form. For a dataset $(X, y^*)$ and cluster assignments $y$, which associates a predicted label $y_i$ to every point $x_i$ in $X$, FMI is defined as follows:
\begin{equation}
    FMI = \frac{TP}{\sqrt{(TP + FP) (TP + FN)}},
\end{equation}
where $TP$, $FP$ and $FN$ respectively represent the number of true positive, false positive and false negative pairs between $y$ and $y^*$. Then, its local form FM$_i$ represents the Fowlkes-Mallows score of datapoint $x_i$ and is defined by:
\begin{equation}
    FMI_i = \frac{TP_i}{\sqrt{(TP_i + FP_i) (TP_i + FN_i)}},
\end{equation}
where $TP_i$, $FP_i$ and $FN_i$ respectively represent the number of true positive, false positive and true negative pairs containing $x_i$. FM$_i$ ranges between $0$ and $1$ and is high if $x_i$ is well clustered with respect to $y^*$. 
From this definition of FM$_i$, we introduce the concept of \emph{FM score per class}:
\begin{equation}
    FM_{C_k} = \frac{1}{N_{C_k}} \sum_{p = 1}^{N_{C_k}} FM_p,
\end{equation}
where $C_k$ represents true class $k$ and $N_{C_k}$ is the number of elements of $C_k$ in the dataset.

\begin{figure}
    \centering
    
\begin{subfigure}[b]{0.48\textwidth}
\begin{center}
\scalebox{0.9}{
\begin{tabular}{ccccc|c}
& NMI & PUR & ACC & FM & FM$_{C_4}$ \tabularnewline \hline
InceptionResnet & \textbf{0.775} & \textbf{0.642} & \textbf{0.588} & \textbf{0.537} & \textcolor{gray}{0.442} \tabularnewline
VGG16 & \textcolor{gray}{0.689} & \textcolor{gray}{0.550} & \textcolor{gray}{0.447} & \textcolor{gray}{0.372} & \textbf{0.653} \tabularnewline
Densenet121 & \textcolor{gray}{0.684} & \textcolor{gray}{0.553} & \textcolor{gray}{0.515} & \textcolor{gray}{0.384} & \textbf{0.700}
\end{tabular}}
\end{center}
\vspace{10pt}
\end{subfigure}

\begin{subfigure}[b]{0.15\textwidth}
\centering \includegraphics[width=\textwidth]{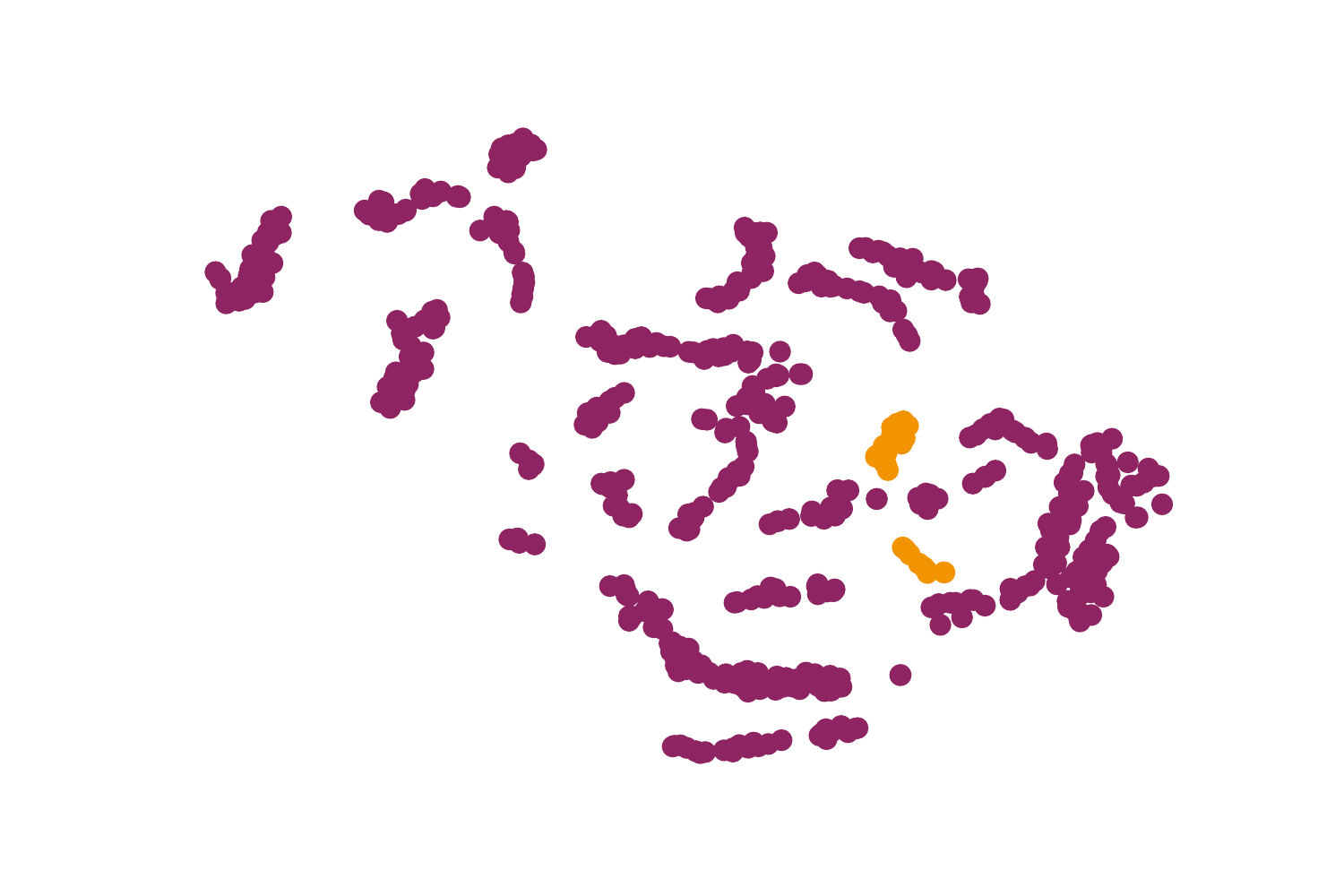}
\caption{InceptionResnet}\label{fig:mvtc_intuition_incRes}
\end{subfigure}
~
\begin{subfigure}[b]{0.15\textwidth}
\centering \includegraphics[width=\textwidth]{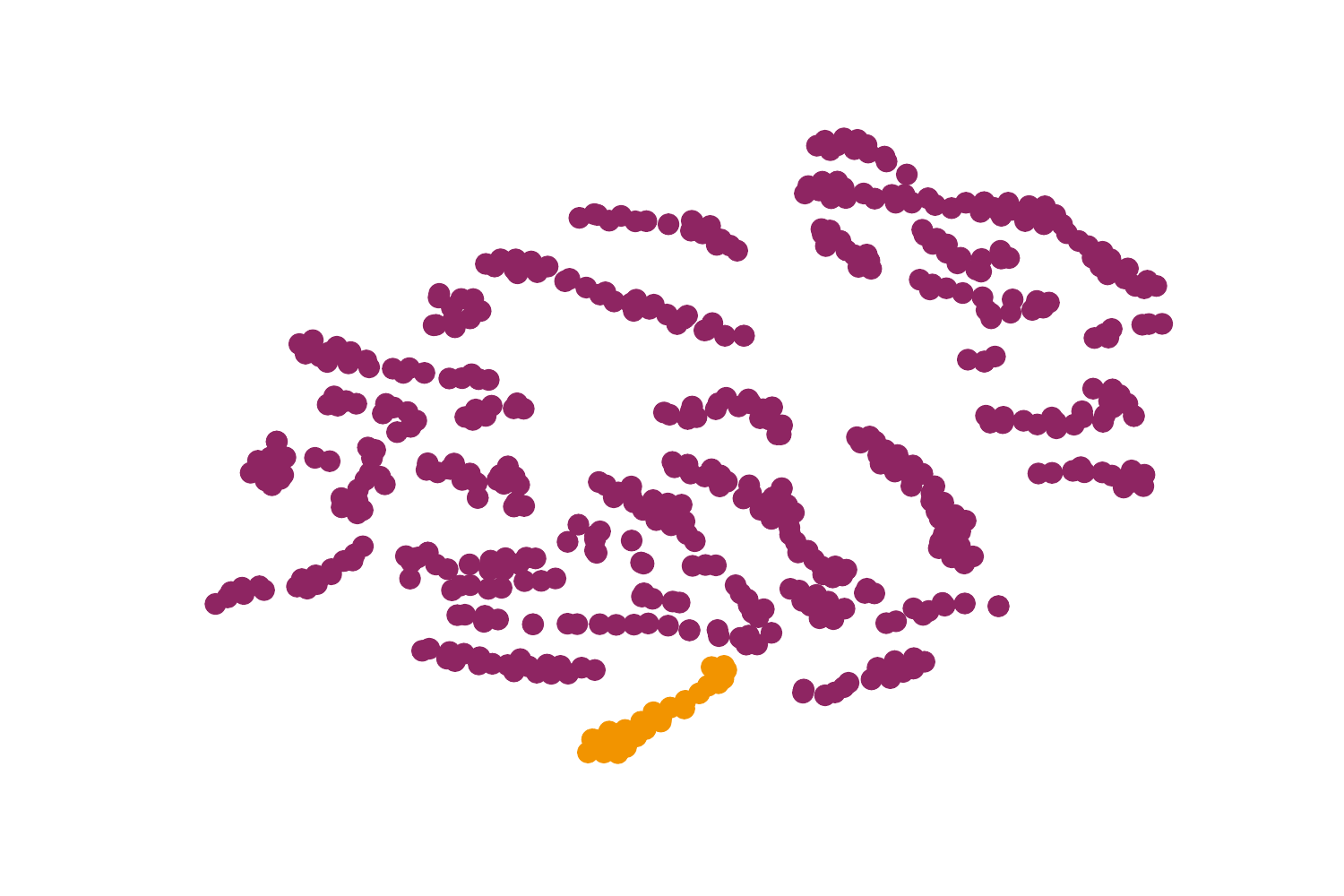}
\caption{VGG16}\label{fig:mvtc_intuition_vgg}
\end{subfigure}
~
\begin{subfigure}[b]{0.15\textwidth}
\centering \includegraphics[width=\textwidth]{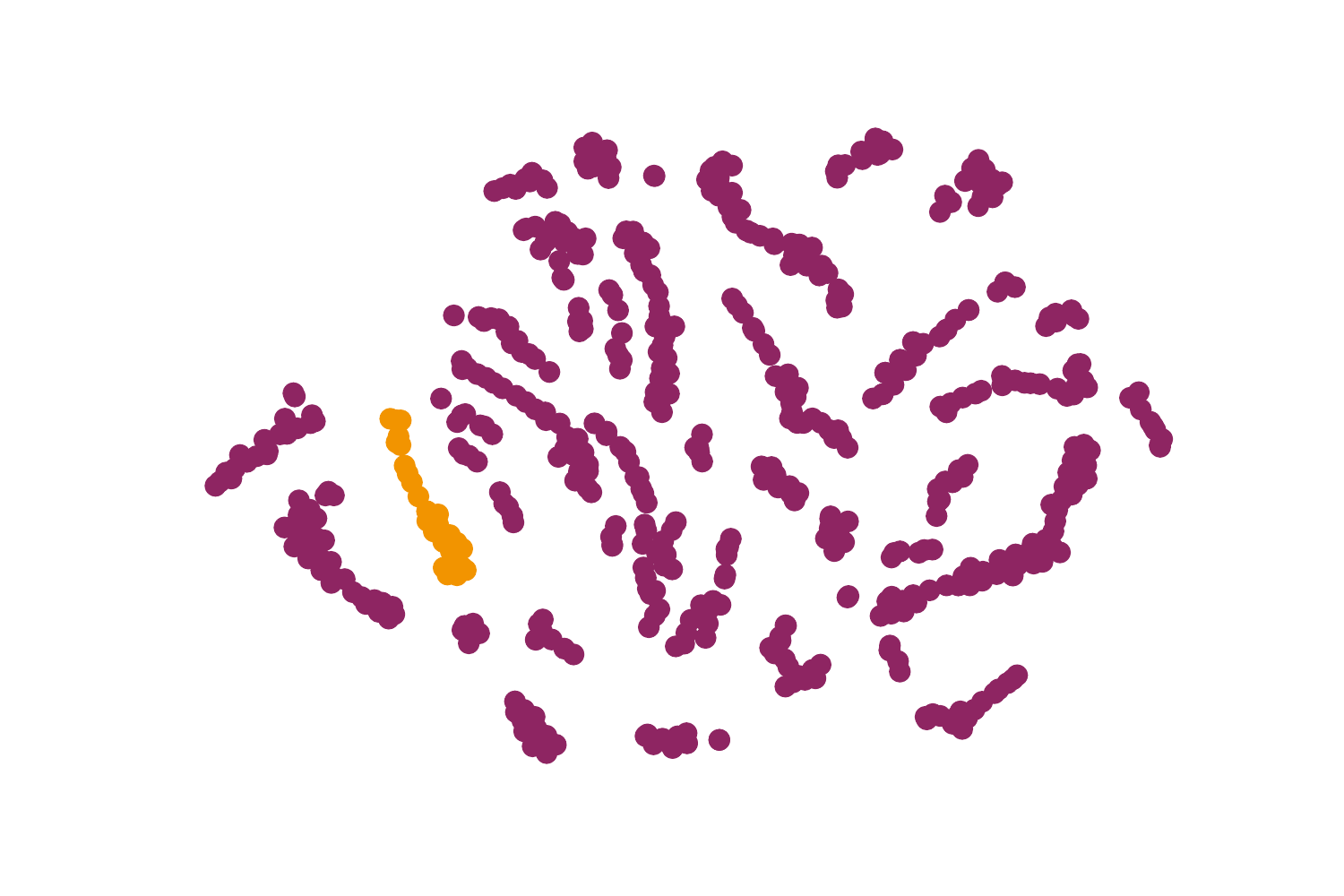}
\caption{Densenet121}\label{fig:mvtc_intuition_dense}
\end{subfigure}

\caption{2d t-SNE  visualization of features extracted by three pretrained CNNs for the UMist dataset. Members of class 4 are in orange. These features form different \emph{complementary} views of the data.}
\label{fig:mvtc_intuition}
\end{figure}

Then, we demonstrate the complementarity of deep feature extractors by carrying out experiments on the UMist dataset. We apply agglomerative clustering to the best performing network on this dataset (InceptionResnet) as well as the two worst performing networks (VGG16 and Densenet121). These networks are introduced in Section~\ref{sec:ic2mvc_expe}. Then, the NMI, PUR, ACC, FM and FM$_{C_4}$ scores are computed. As shown in the table in Figure~\ref{fig:mvtc_intuition}, InceptionResnet is performing way better than its two competitors with respect to all global metrics. However, looking at class 4, we can see that the two other networks present a significant improvement over InceptionResnet. The 2d t-SNE \cite{tsne} representations of the features extracted with the different CNNs are also represented in Figure~\ref{fig:mvtc_intuition}. Members of class 4 are in orange and the other classes in purple. For VGG16 and Densenet121, the feature representations of class 4 are more compact and isolated, which explains why they perform better on this class. This experiment demonstrates the complementarity of the different networks on one example and justifies the proposed multi-view clustering approach.

\begin{figure*}[!ht]
\centering

\includegraphics[width=0.75\textwidth]{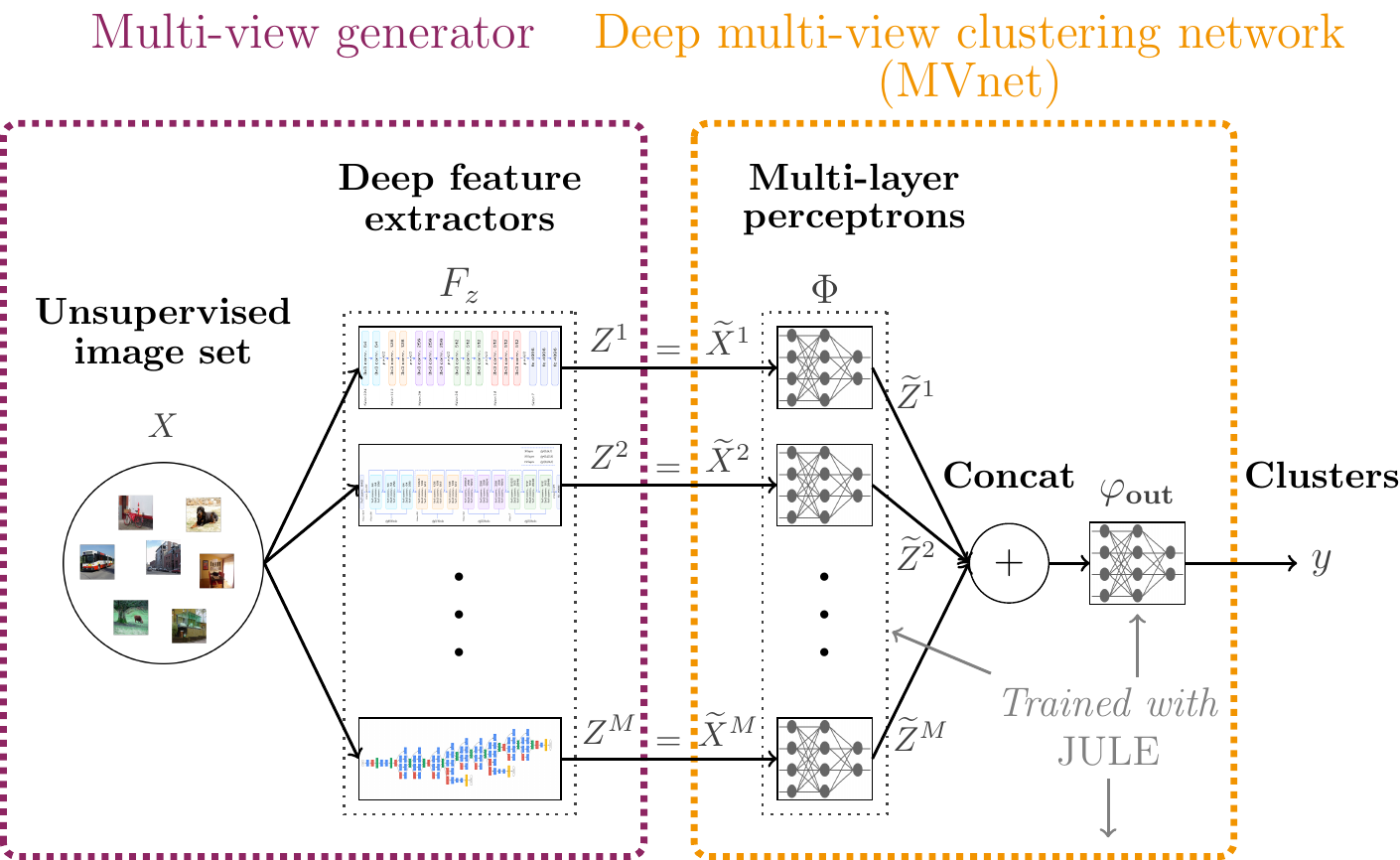}

\caption{Proposed multi-view generation + deep multi-view clustering (DMVC) approach to solve the Image Clustering problem.}
\label{fig:dmvc}
\end{figure*}

\subsection{IC problem reformulation}
\label{sec:ic2mvc_formulation}

Let $X = \{x_1, ..., x_{N}\} \subset \mathcal{I}$ be an unlabeled set of $N$ natural images, and let $F_z = \{f_z^1, ... f_z^M\}$ be a set of $M$ feature extractors. In theory, $F_z$ can be composed of any function mapping raw pixel representations to lower-dimensional vectors, but in practice, we use pretrained deep CNNs. The first step in our approach is to generate a set of feature vectors from each element of $F_z$. For all $i \in [1, ... M]$, we denote $Z^j$ the matrix of features representing $X$ such that, its row $Z^j_i$ is the feature vector representing $x_i$ and extracted by $f_z^j$:
\begin{equation}
\label{eq:feat}
    Z^j_i = f_z^j(x_i).
\end{equation}
In other words, $\text{Z} = \{Z^1, ..., Z^M\}$ can be interpreted as a set of views representing $X$. Thus, $\text{Z}$ is a multiview dataset representing $X$ and the problem of clustering $\text{Z}$ is a MVC problem, which can be solved using any MVC algorithm~\cite{survey_mvc}. A visual representation of the multi-view generation mechanism can be seen in the purple frame of Figure~\ref{fig:dmvc}.

\vspace{0.1\baselineskip}
\noindent\textit{\underline{Remark:} All along this section, we use letter $i$ for indexing across data samples, letter $j$ for indexing across feature extractors and letter $k$ for indexing across clusters. Similarly, $N$, $M$ and $K$ respectively stand for the number of data samples, feature extractors and clusters.}

\subsection{Experimental evidences}
\label{sec:ic2mvc_expe}

To conclude this section on artificial multiview data generation
, we give first experimental evidences and study the optimal number of networks to use. To do so, experiments are carried out on 4 of the standard datasets introduced in Section~\ref{sec:benchmark_exp_design_dataset}: VOC2007, Archi, Flowers and UMist. In these experiments, as well as in the rest of the section, we use 10 pretrained architectures: VGG16, VGG19, Inception, Xception, Resnet50, Densenet121, Densenet169, Densenet201~\cite{densenet}, Nasnet~\cite{nasnet} and InceptionResnet~\cite{inceptionResnet}. In this section, MVC problems are solved using the Multi-View Ensemble Clustering method~\cite{multiview_ensemble} with agglomerative clustering (MVEC$_{\text{agg}}$). This method consists in clustering each view separately using agglomerative clustering and generating a consensus partition using an ensemble clustering method based on co-association matrix. For more information about this method, the reader can refer to~\cite{ensemble_survey}. We choose agglomerative clustering as our base algorithm because its simplicity enables us to study straightforwardly the clusterability of the different feature spaces. Agglomerative clustering is also preferred over K-means because it does not depend on initialization and thus avoids random effects in the results. 

\begin{figure*}[t]
\centering

\includegraphics[width=0.95\textwidth]{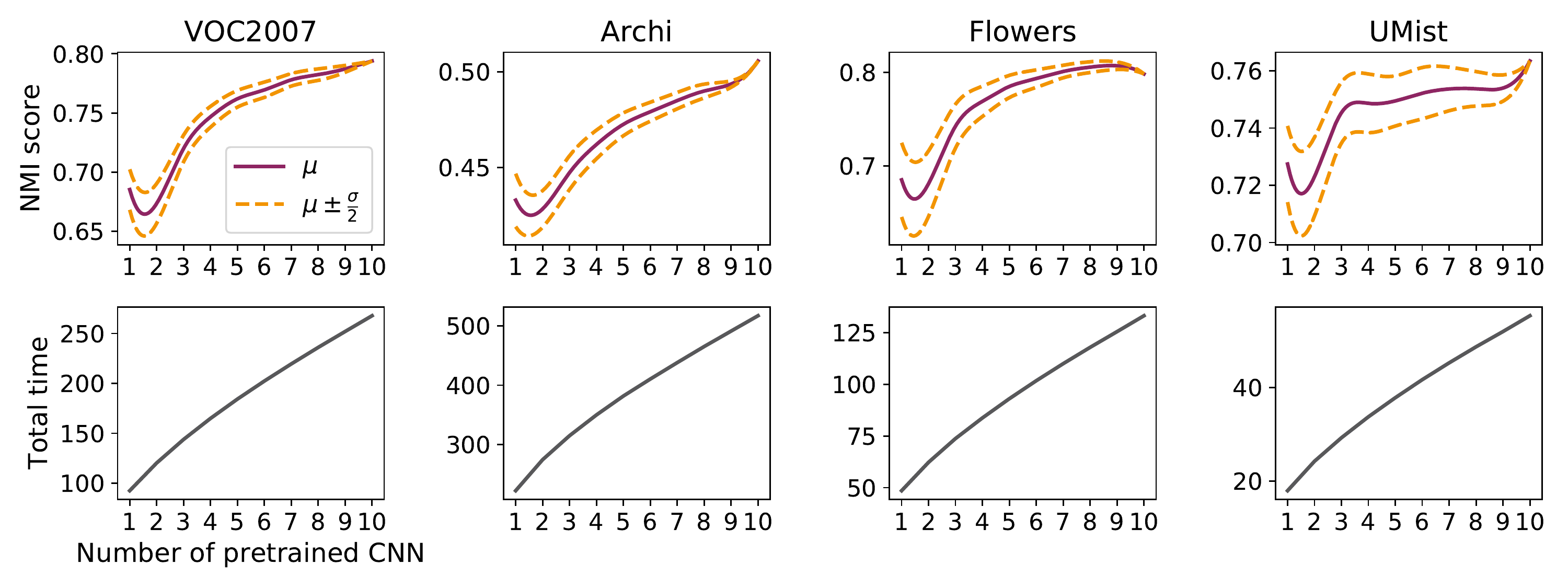}

\caption[Influence of the number of CNN feature extractors]{Evolution of the NMI score and total time (in sec) for different numbers of pretrained CNN feature extractors.}
\label{fig:mvtc_expeMV}
\end{figure*}

Then, to study the clustering quality of multiview data from $m~(\leq 10)$ CNNs, we generate all the multiview problems from all the possible combinations among the different architectures. Each of these problems is then solved with MVEC$_{\text{agg}}$ and the NMI scores are computed. In Figure~\ref{fig:mvtc_expeMV}, for all $m \in \{1, ..., 10\}$ we report the mean and standard deviation across the $C_m^{10}$ NMI scores. To generate Figure~\ref{fig:mvtc_expeMV}, for each dataset, we need to solve $\sum\limits_{m=1}^{10} C_m^{10} = 1023$ MVC problems. These results suggest that combining more networks both increases the clustering accuracy on average and decreases the variability, which can be seen as the risk to obtain poor results. These results are in line with the intuitions from Section~\ref{sec:ic2mvc_intuition}. From these simple experiments, we decide to use all ten networks in the rest of this paper, which is likely to give the most robust clustering results. It is also interesting to note that the two networks case performs worst than the single network case for all four datasets. This probably comes from the absence of a ``majority'' to distinguish which information is relevant.

\subsection{Considerations on time complexity}
\label{sec:ic2mvc_expe_time}

The evolution of the clustering time with the number of feature extractors is also reported in Figure~\ref{fig:mvtc_expeMV}. All the experiments were conducted on the same machine. Obviously, when more networks are added to the pipeline, the total clustering time increases. However, when running MVEC$_{\text{agg}}$ with $m$ networks, if we have $m$ GPUs available, each feature extraction and agglomerative clustering can be run in parallel on a dedicated GPU. Thus, for a given dataset, if we note
\begin{itemize}
    \item $t_1^j$ the time for feature extraction with $f_z^j$,
    \item $t_2^j$ the time for running agglomerative clustering on $Z^j$ and
    \item $t_3$ the time for merging the different partitions into a consensus partition,
\end{itemize}
the total clustering time is $\max\limits_{j}~\left(t_1^j + t_2^j\right) + t_3$. For this reason, the time for running MVEC$_{\text{agg}}$ does not increase linearly with the number of feature extractors. Using 10 networks instead of one only increases the clustering time by a factor of around 2.5, As illustrated on Figure~\ref{fig:mvtc_expeMV}. The increase in the total time comes from the fact that the slowest networks are selected more often when more networks are used and that the partitions grouping becomes slower. In practice the slowest step is feature extraction. If only $m\prime (< m)$ GPUs are available, then the total time increases by a factor of $\sim \left\lceil\frac{m}{m\prime}\right\rceil$, where $\left\lceil.\right\rceil$ denotes the ceiling function. Making the method scalable to a large number of CNNs when few GPUs are available is a potential area of improvement for the method.

\section{Deep Multi-View Clustering}
\label{sec:dmvc}

\subsection{Deep multi-view clustering (DMVC)}
\label{sec:dmvc_method}

In this section, we define our approach for solving MVC using end-to-end clustering. A deep clustering framework is defined by a loss function $\mathcal{L}$ and a procedure $\mathcal{P}$ to optimize the loss function. Multiple approaches have already been adopted to define the clustering-oriented loss $\mathcal{L}$ and the optimization procedure $\mathcal{P}$. 
Let $\widetilde{X}$ be an unsupervised dataset \footnote{The proposed solution to Image clustering (IC) is in two steps: feature extraction and deep clustering. The ``tilde'' notation is introduced to avoid confusion between the different input spaces and latent spaces. Hence, $X$ and $Z$ refer respectively to the input and latent spaces of the deep feature extractor $f_z$, which weights are fixed, whereas $\widetilde{X}$ and $\widetilde{Z}$ refer to the deep clustering network $\varphi_{\theta}$, parameterized by $\theta$. Obviously, if features from a single feature extractor are used to train a single deep clustering network, we have $Z = \widetilde{X}$.}. To solve $\widetilde{X}$ using deep clustering, one first needs to specify a neural network architecture $\varphi_{\theta}$, parameterized by $\theta$, which projects $\widetilde{X}$ into a lower dimensional feature space: $\widetilde{Z}_{\theta} = \varphi_{\theta}(\widetilde{X})$. Then, $\mathcal{P}$ is applied to minimize $\mathcal{L}(\theta, \widetilde{X})$, producing both a good representation $\widetilde{Z}_{\theta_{\text{final}}}$ and a set of cluster assignments $y_{\text{final}}$.

The choice of the architecture of $\varphi_{\theta}$ usually depends on the kind of dataset to solve. For example, when dealing with large images, $\varphi_{\theta}$ can be a CNN and when $\widetilde{X}$ is composed of smaller vectors, $\varphi_{\theta}$ can be a multi-layer perceptron (MLP). In the case of MVC, each element of $\widetilde{X}$ is a collection of vectors. For example, the $i^{th}$ element of $\widetilde{X}$ is written as $\widetilde{X}_i = \{Z^j_i; \forall j \in [1, ... M]\}$. For this reason, to embed MVC into a deep clustering framework, we need to define a multi-input neural network architecture for $\varphi_{\theta}$, which we call MVnet. MVnet consists of a set of $M$ independent MLPs, denoted $\Phi = \{\varphi_{\theta_1}, ... \varphi_{\theta_M}\}$, such that, $\forall j \in [1, ... M]$, the dimension of the input layer of $\varphi_{\theta_j}$ is equal to the dimension of the output layer of the associated $f_z^j$. We also define $\varphi_{\theta_{out}}$, another MLP with input layer dimension equal to the sum of the dimensions of the output layers over the elements of $\Phi$. Thus, an MVnet is composed of three layers: a parallel layer containing all the elements of $\Phi$, followed by a concatenating layer that feeds into $\varphi_{\theta_{out}}$. A visual representation of the MVnet architecture can be seen in the orange box in Figure~\ref{fig:dmvc}. We note that all the elements of $\Phi$ are independent and do not share any weights.

DMVC is a generic framework and MVnet can be optimized using most deep end-to-end clustering approaches. In practice, we have tried to implement DMVC within both the Improved DEC framework \cite{idec}, and the JULE framework \cite{jule}. After carrying out experiments on standard datasets, we noticed that JULE performs significantly better \cite{bmvc18}. In addition, implementing IDEC on a new dataset requires additional parameters tuning to pretrain the autoencoders, which is time consuming, potentially error-prone and less generic. For these two reasons, in the rest of this paper, we adopt JULE to solve the DMVC problem. The proposed JULE-DMVC implementation is explained in more details in the next section.



\subsection{DMVC with JULE}
\label{sec:dmvc_jule}

Joint Unsupervised Learning of Deep Representations and Image Clusters, or JULE, is an iterative end-to-end clustering process that has demonstrated excellent experimental results on several natural image datasets. In this section, we propose to leverage JULE to train an MVnet to solve the MVC problem. We start by presenting an overview of the standard JULE framework and then propose an extention to adapt it to MVnet. 
For a more complete description of JULE, we refer the reader to the original paper \cite{jule}.

\subsubsection{JULE overview}
\label{sec:dmvc_jule_jule}

Let $\widetilde{X}$ be an unsupervised dataset 
that we aim to cluster into $K^*$ groups. 
Let $\varphi_{\theta}$ denote a neural network parameterized by $\theta$, which produces a lower dimensional representation of the initial dataset $\widetilde{Z}_{\theta} = \varphi_{\theta}(\widetilde{X})$. JULE is an iterative process, hence we introduce $\theta[t]$ and $y[t]$, the values of the weights and cluster assignments at iteration $t$. At step $t$, the cluster assignment $y[t]$ defines a set of $K[t] \leq K^*$ clusters and $\theta[t]$ defines a latent space $\widetilde{Z}_{\theta[t]}$.

JULE is an iterative optimization process which leverages alternating optimization to obtain both good cluster assignments $y[t_f]$ and a good new latent representation of the initial data $\widetilde{Z}_{\theta[t_f]}$, where $t_f$ represents the state of the clustering model after optimization. Going from iteration $t: (\theta[t], y[t])$ to iteration $t+1$ consists in solving two subproblems:
\begin{itemize}
\item Representation learning: The initial network $\varphi_{\theta[t]}$ is trained on the dataset $(\widetilde{X}, y[t])$ to generate a set of updated weights $\theta[t+1]$.

\item Clusters merging: A new set of cluster assignments $y[t+1]$ are generated from similarities computed in $\widetilde{Z}_{\theta[t+1]}$.
\end{itemize}
Because clusters are being merged, the total number of clusters decreases when we progress through the optimization: $t < t^{\prime} \Rightarrow K[t] < K[t^{\prime}]$. JULE stops at iteration $t_{f}$ such that $K[t_f] = K^*$. As a final step, the network $\varphi_{\theta[t_f]}$ can optionally be trained on $y[t_f]$ to fine-tune the representation. 


Although the precise method for cluster merging and representation learning is not detailed here, it is worth saying a few words about initialization as it differs for multi-view data. The initial set of clusters $y[t_0]$ is computed using the initialization method proposed in \cite{jule_init}. At first, a cluster is created for each sample, containing the sample and its nearest neighbor in the input space $\widetilde{X}$, thus creating $N$ clusters. Then, the number of clusters is reduced by merging clusters which contains duplicated samples. This heuristic process usually lead to clusters which contain between 3 to 5 samples. The weights of the neural network $\theta[t_0]$ are initialized using Xavier initialization~\cite{xavier}.

\subsubsection{JULE with multiview data}
\label{sec:dmvc_jule_mv}

In this section, we propose to use JULE to train an MVnet to solve the MVC problem. The initialization step for JULE requires to merge the first clusters based on distances in the initial feature space of the data. To avoid defining a distance in a multi-view space, a different approach is adopted. First, each $\varphi_{\theta_j}$ is pretrained separately on $Z^j$. Then, $\varphi_{\theta_{out}}$ is trained on the concatenation of the $\widetilde{Z}^j = \varphi_{\theta_j}(Z^j)$. Once MVnet has been properly initialized, it is used to produce a meaningful initial unified latent representation of the multiview data. This representation serves as the initial space in which the first cluster labels are assigned. Once the first clusters are initialized, JULE can be carried out normally on the MV data.

Another straightforward way to use JULE to solve MVC is to concatenate the different views and apply JULE to the concatenated features. This method is taken as a baseline for comparison in order to evaluate the MVnet approach and is referred to as the Concatenate and Cluster approach (CC).

\section{Experimental validation}\label{sec:exp_mvc}

\subsection{Experimental setup}
\label{sec:exp_mvc_setup}

Our experiments are conducted on the same 8 datasets presented in Section~\ref{sec:benchmark_exp_design_dataset}. In addition, we also add CIFAR10~\cite{cifar10} to the list of datasets for evaluation. CIFAR10 is not particularly well suited to the feature extraction process presented here as the original images are only 32x32 pixels. This implies that the images need to be resized to ten times their original sizes, which generates very noisy input images for the feature extractors. However, we believe that it is interesting to see the behavior of our approach on such a dataset.

For multiview generation, we use the Keras implementations and pretrained weights of the ten CNN architectures introduced in Section~\ref{sec:ic2mvc_expe}. For each network, the chosen layer is the last before softmax, as suggested by the experimental results of Section~\ref{sec:benchmark_results_layer}.

To solve the generated MVC problem, we implement the two proposed DMVC methods (CC and MVnet). 
For MVnet, we also report the results without fine tuning (MVnet$_{\text{fix}}$), i.e. just after the initialization of each MLP. In order to have a standard baseline for comparison, the results are compared to MVEC with agglomerative clustering (MVEC), Multi-View Spectral Clustering~\cite{mvsc} (MVSC) and Robust Multiple Kernel K-Means~\cite{rmkkm_used} (RMKKM). These methods are standard approaches to deal with multi-view data and have demonstrated good results on different MVC datasets. For completeness, the JULE variants of these three methods are also implemented. It consists in using JULE clustering instead of agglomerative for MVEC$_{\text{jule}}$, and in applying the MVC algorithms on the feature representations obtained after applying JULE to each individual feature extractor for MVSC$_{\text{jule}}$ and RMKKM$_{\text{jule}}$.

DMVC is a framework for \emph{unsupervised} classification, hence, hyperparameter tuning should be avoided. In all of our experiments, we use default parameters for every sub-algorithm used. For agglomerative clustering, we use the default configuration of the scikit-learn implementation. For MVEC, the co-association matrix is clustered with agglomerative clustering with average linkage. For both MVSD and RMKKM, Gaussian Radial Basis Function kernels are used and the sigma parameter is set automatically using the Sigest heuristic~\cite{sigest}. Finally, for JULE, we use the hyperparameters recommended in the original paper \cite{jule}. We also use the same kind of neural network architecture used in the original paper:
\begin{itemize}
    \item all the MLPs constituting the MVnet architecture used in our experiments have dimensions $d-160-160$, where $d$ is the input dimension,
    \item the activation functions for the hidden layer are rectified linear unit,
    \item and $l2$-regularization is used during training.
\end{itemize}
The clustering results are evaluated using both normalized mutual information (NMI), purity (PUR) and clustering accuracy (ACC).

\subsection{Experimental results}
\label{sec:exp_mvc_res}

All the results of our experiments can be found in \ref{annex:full_results_mvtc}. They report NMI, PUR and ACC scores for every pretrained CNN independently as well as for the different MVC methods applied to the MVC problems generated with the ten feature extractors. It is difficult to draw conclusions from the large number of results reported in \ref{annex:full_results_mvtc}, hence, this section presents a condensed version of these results.

\begin{figure}[!ht]
\centering

\begin{subfigure}[b]{0.45\textwidth}
\includegraphics[width=\textwidth]{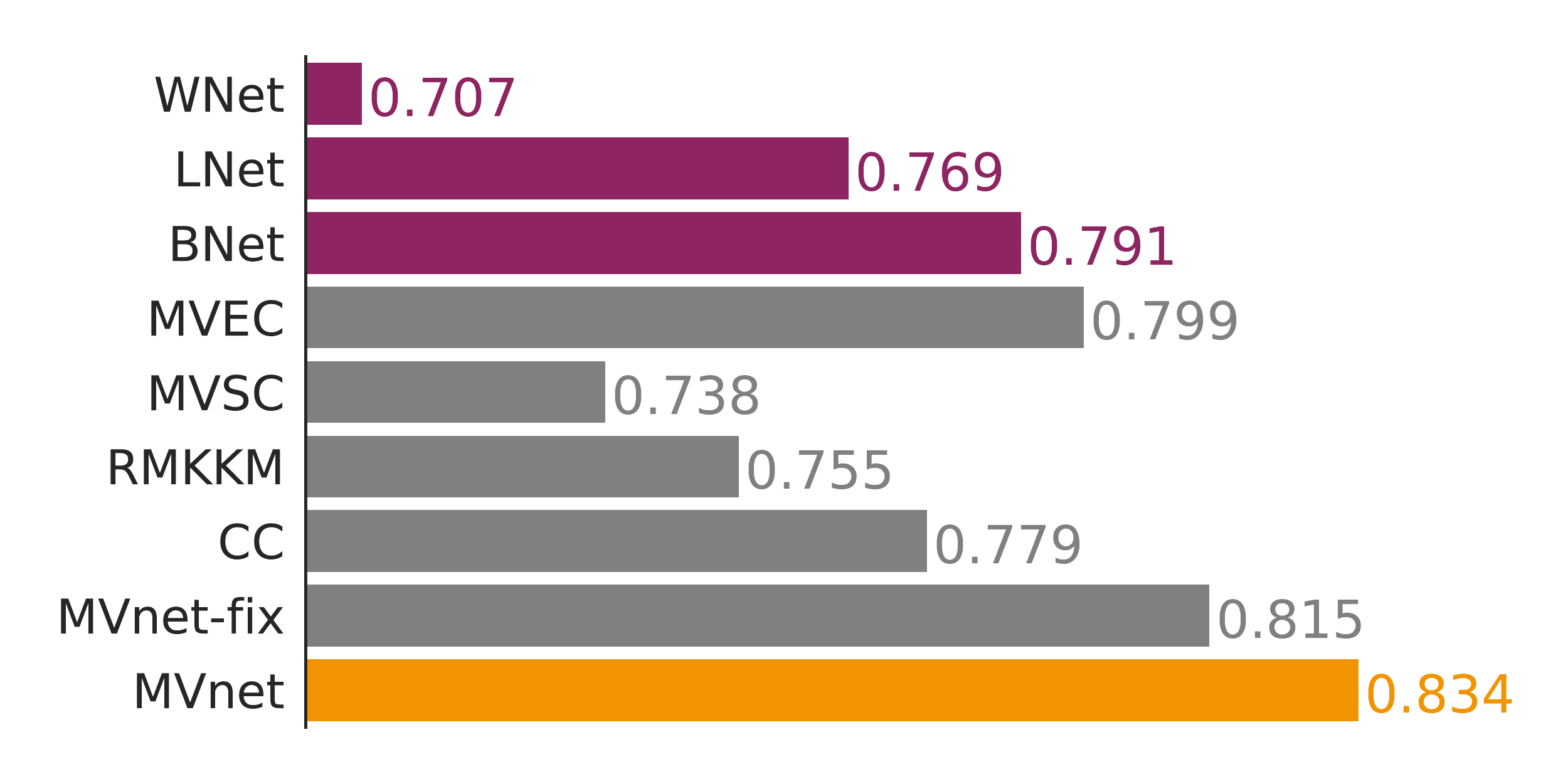}
\vspace{-25pt}
\caption{VOC2007}\label{fig:mvtc_allResults_voc}
\end{subfigure}

\begin{subfigure}[b]{0.45\textwidth}
\includegraphics[width=\textwidth]{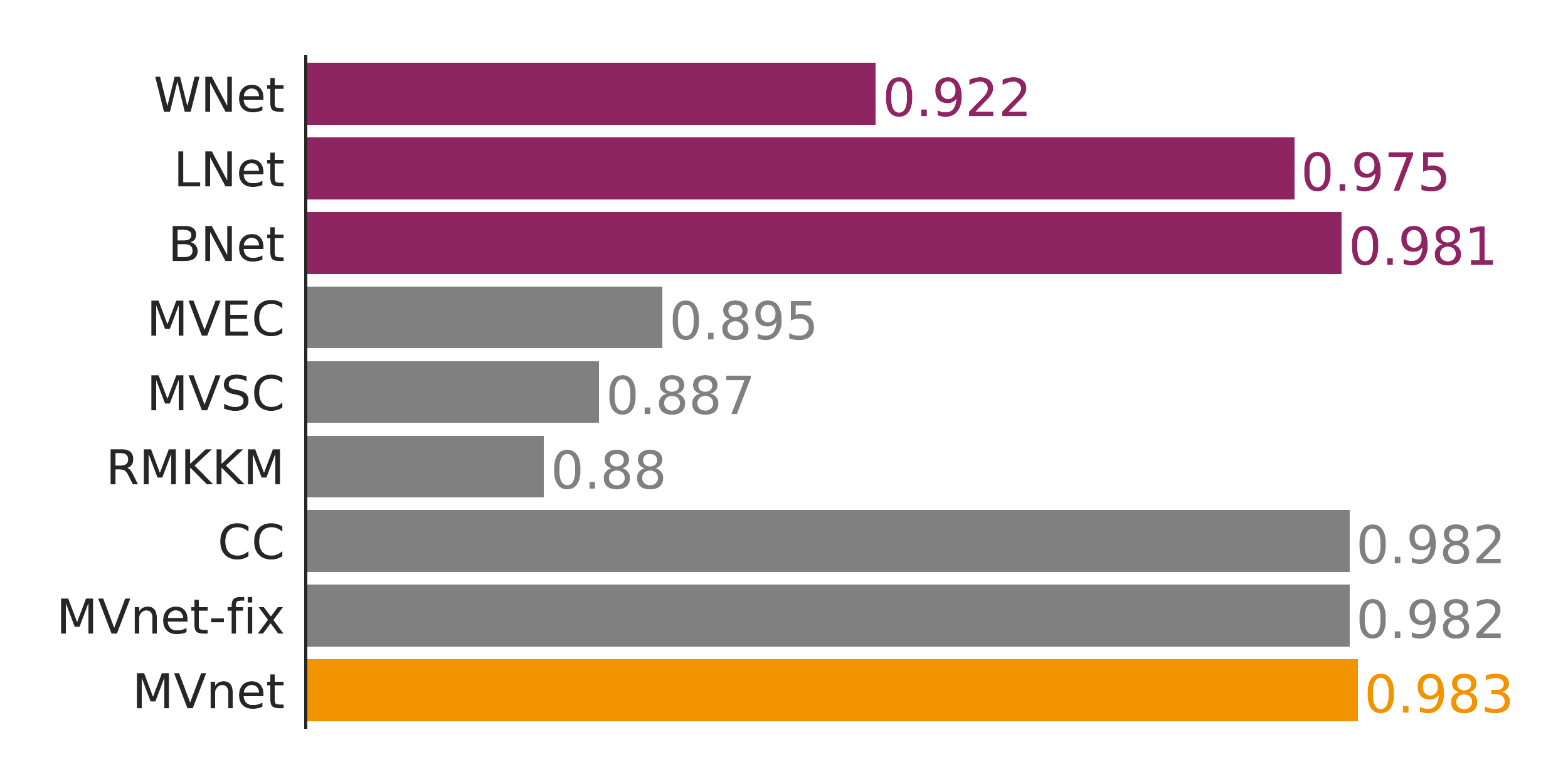}
\vspace{-25pt}
\caption{COIL100}\label{fig:mvtc_allResults_coil}
\end{subfigure}

\begin{subfigure}[b]{0.45\textwidth}
\includegraphics[width=\textwidth]{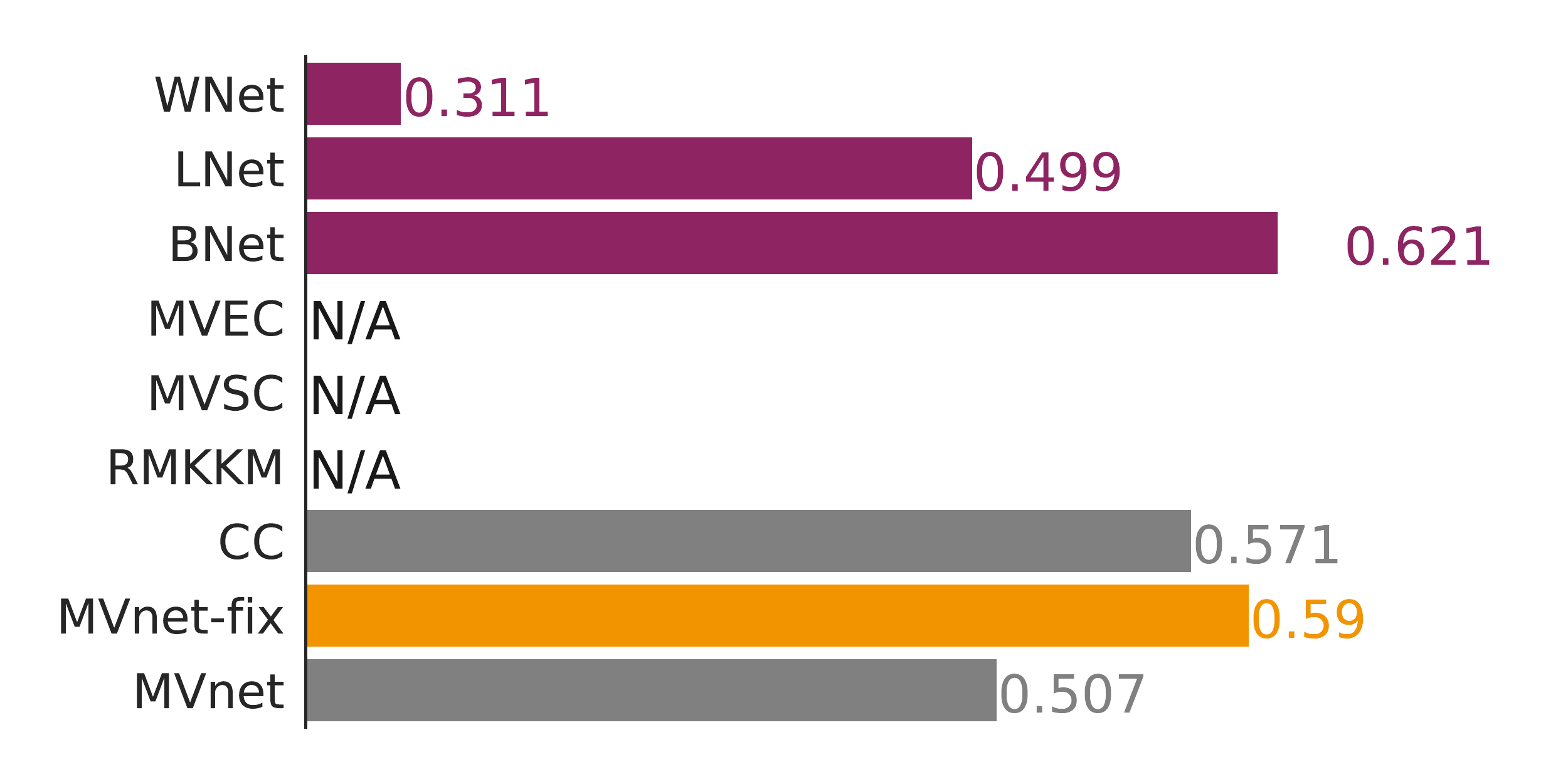}
\vspace{-25pt}
\caption{CIFAR10}\label{fig:mvtc_allResults_cifar}
\end{subfigure}

\caption{MIX score values for the different MVC methods applied to object recognition datasets. \textit{Purple: single extractor,  gray: multi-view methods, orange: best multi-view}.}
\label{fig:mvtc_allResults_object}
\end{figure}

\begin{figure}[!ht]
\centering
\begin{subfigure}[b]{0.45\textwidth}
\includegraphics[width=\textwidth]{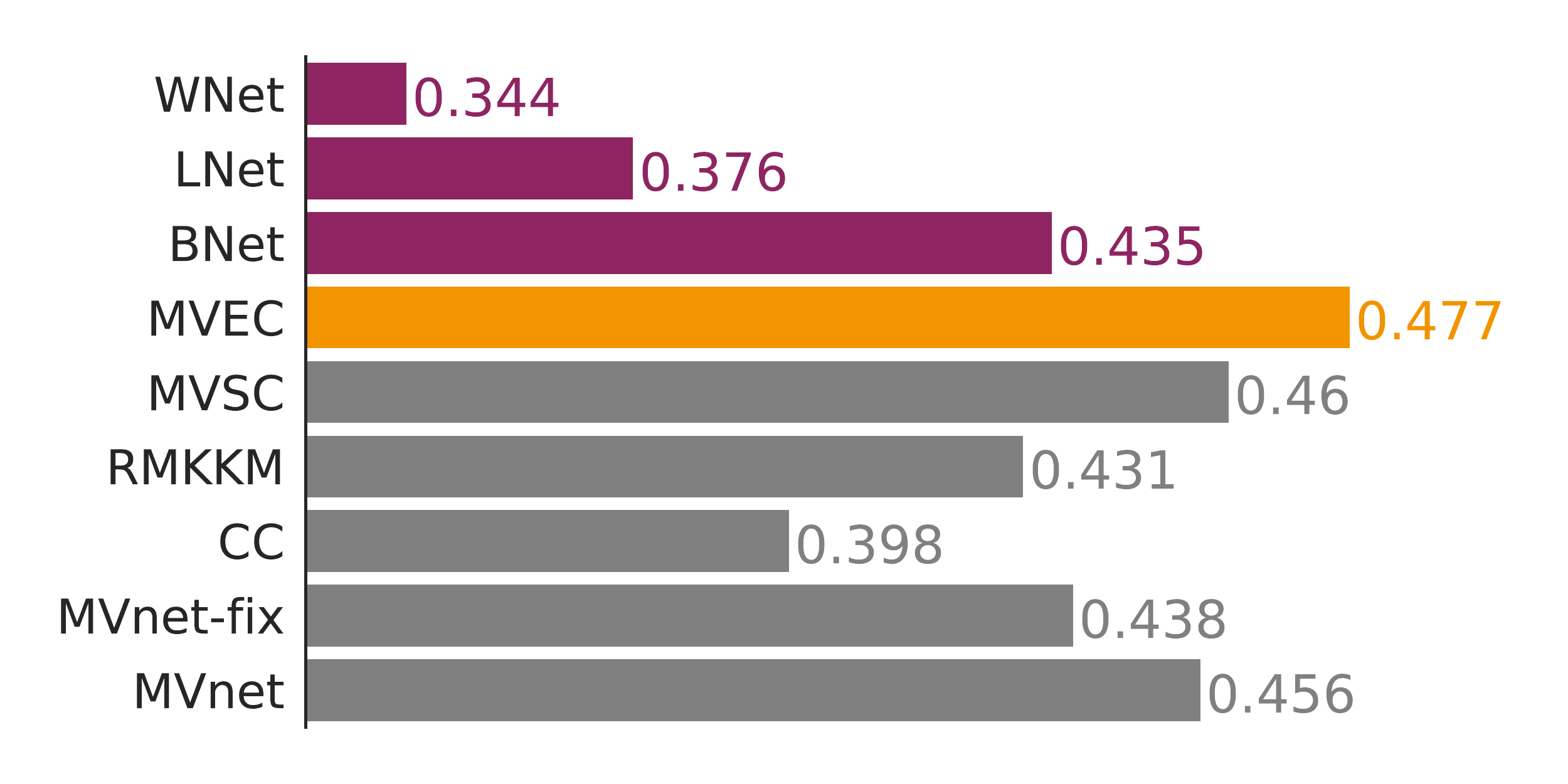}
\vspace{-25pt}
\caption{Archi}\label{fig:mvtc_allResults_archi}
\end{subfigure}

\begin{subfigure}[b]{0.45\textwidth}
\includegraphics[width=\textwidth]{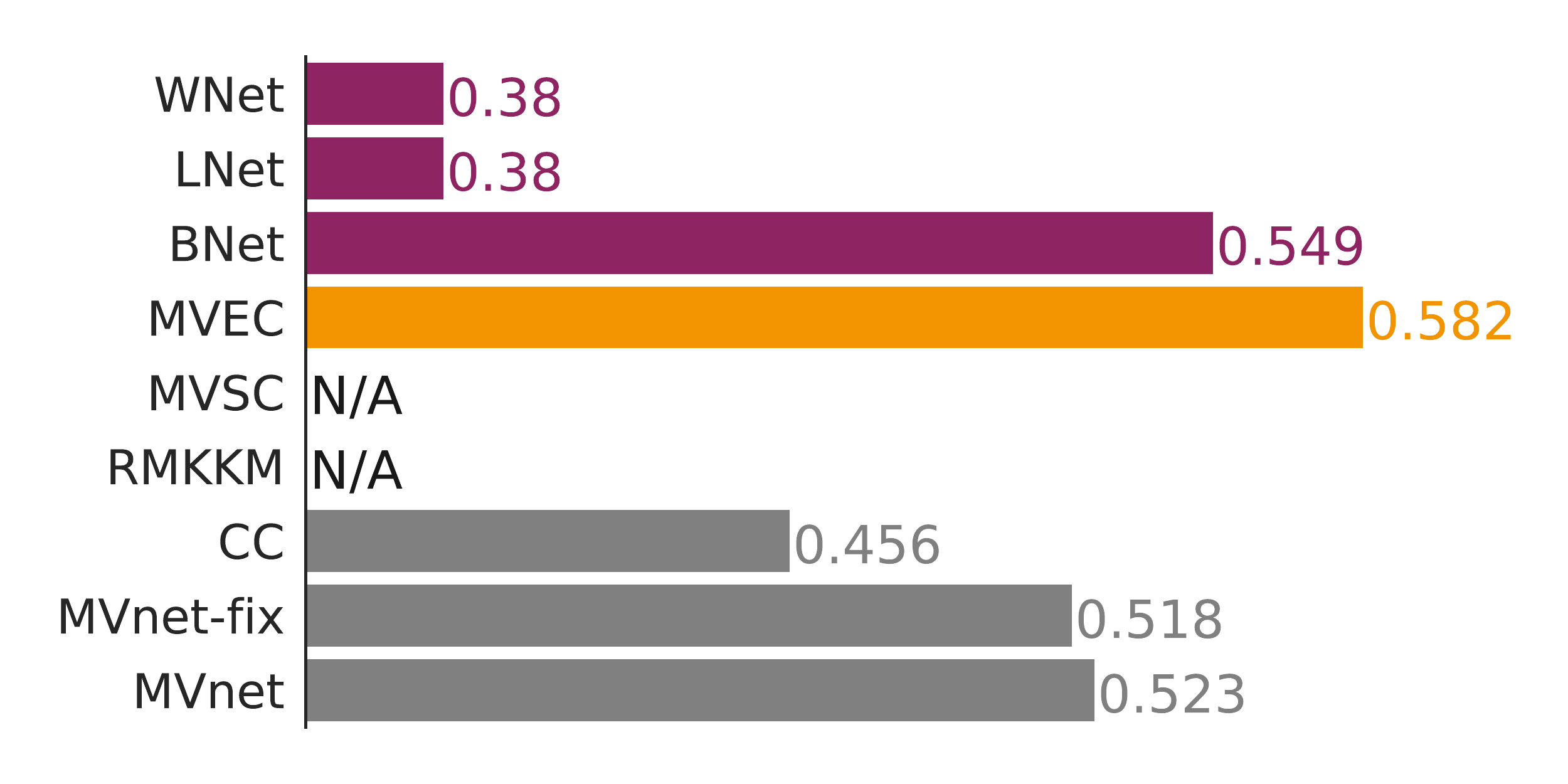}
\vspace{-25pt}
\caption{MIT}\label{fig:mvtc_allResults_mit}
\end{subfigure}

\caption{MIX score values for the different MVC methods applied to scene datasets. \textit{Purple: single extractor,  gray: multi-view methods, orange: best}.}
\label{fig:mvtc_allResults_scene}
\end{figure}

\begin{figure}[!ht]
\centering
\begin{subfigure}[b]{0.45\textwidth}
\includegraphics[width=\textwidth]{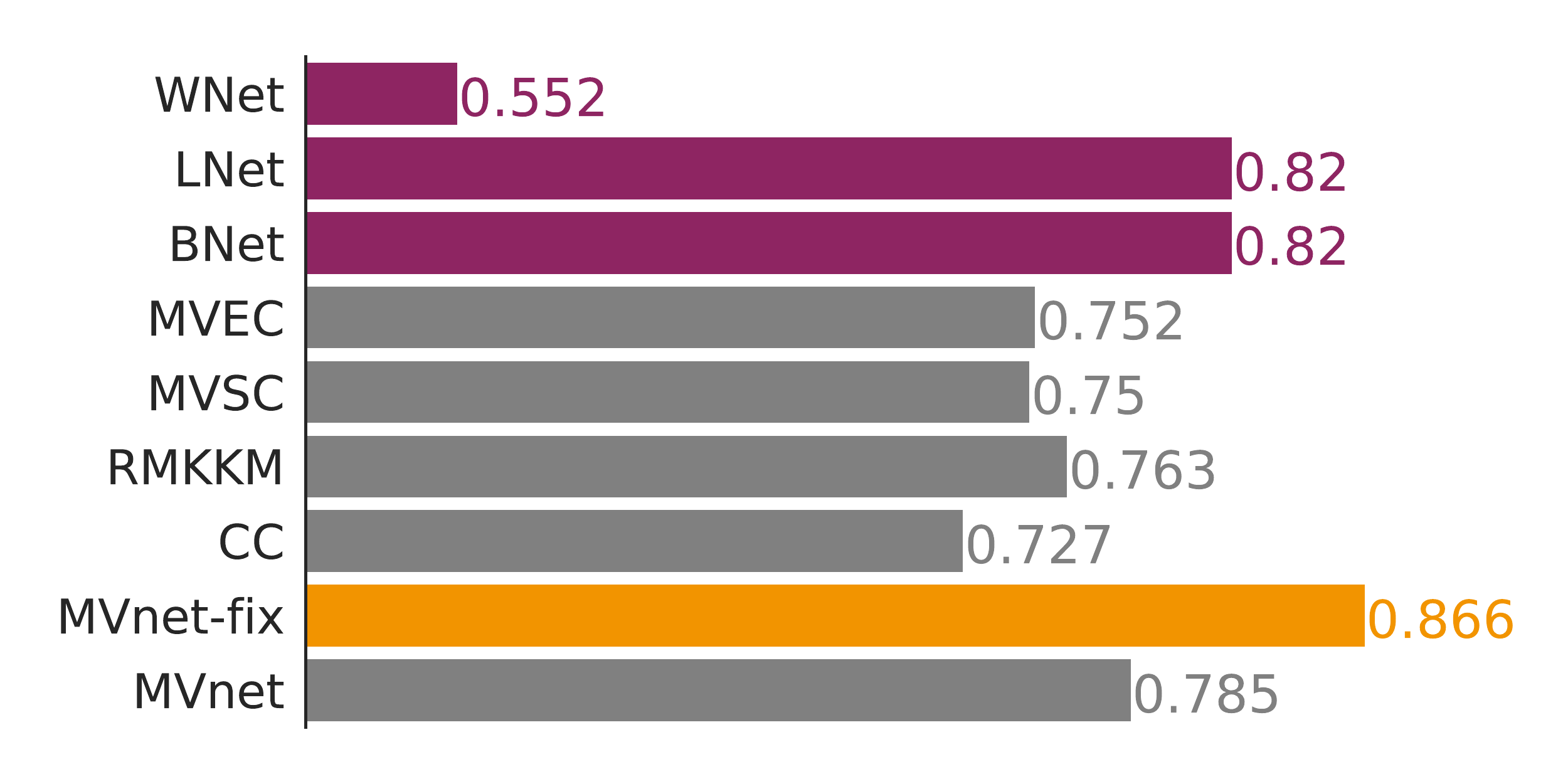}
\vspace{-25pt}
\caption{Flowers}\label{fig:mvtc_allResults_flowers}
\end{subfigure}

\begin{subfigure}[b]{0.45\textwidth}
\includegraphics[width=\textwidth]{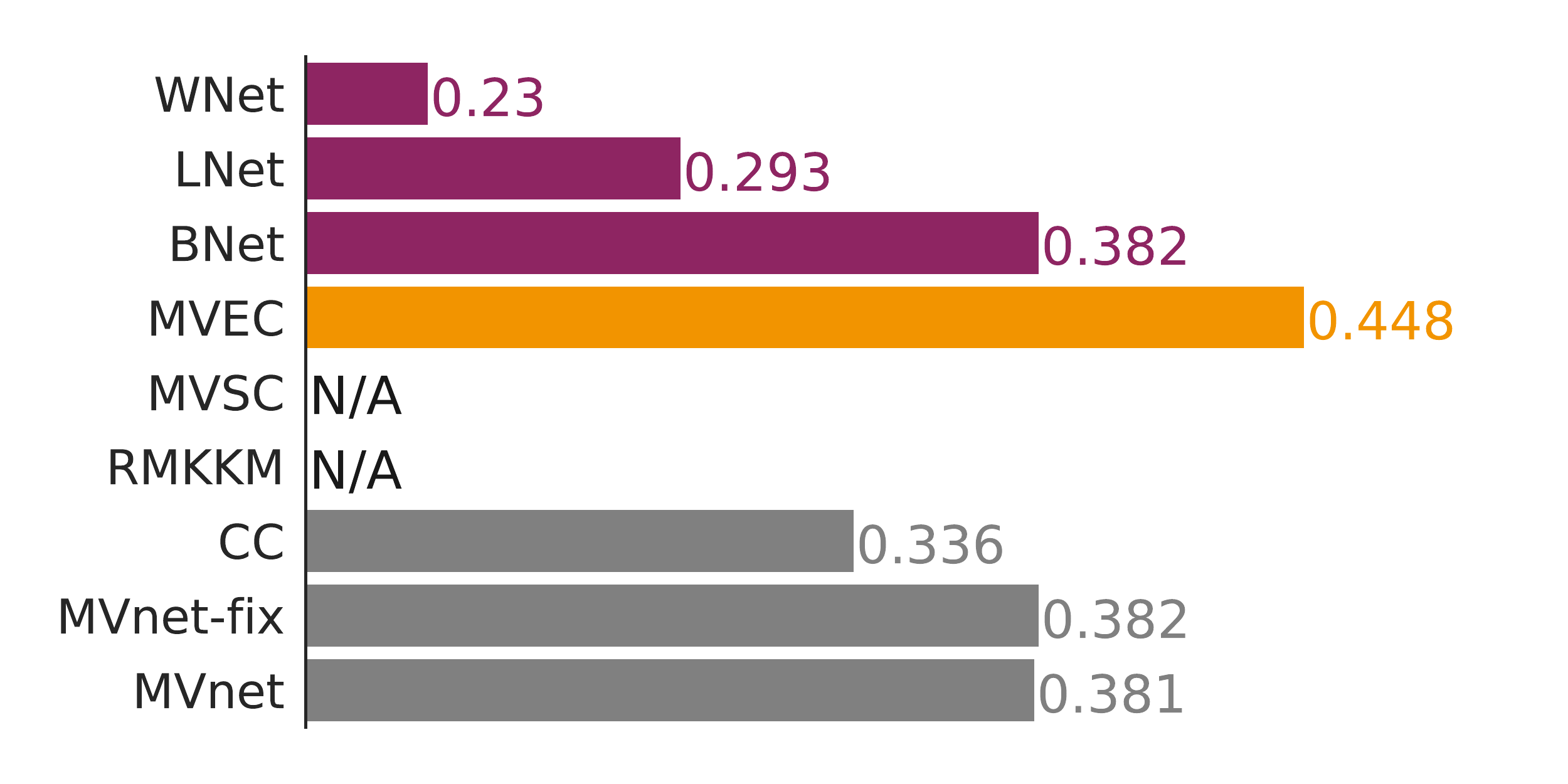}
\vspace{-25pt}
\caption{Birds}\label{fig:mvtc_allResults_birds}
\end{subfigure}

\caption{MIX score values for the different MVC methods applied to fine-grained datasets. \textit{Purple: single extractor,  gray: multi-view methods, orange: best}.}
\label{fig:mvtc_allResults_FG}
\end{figure}

\begin{figure}[!ht]
\centering
\begin{subfigure}[b]{0.45\textwidth}
\includegraphics[width=\textwidth]{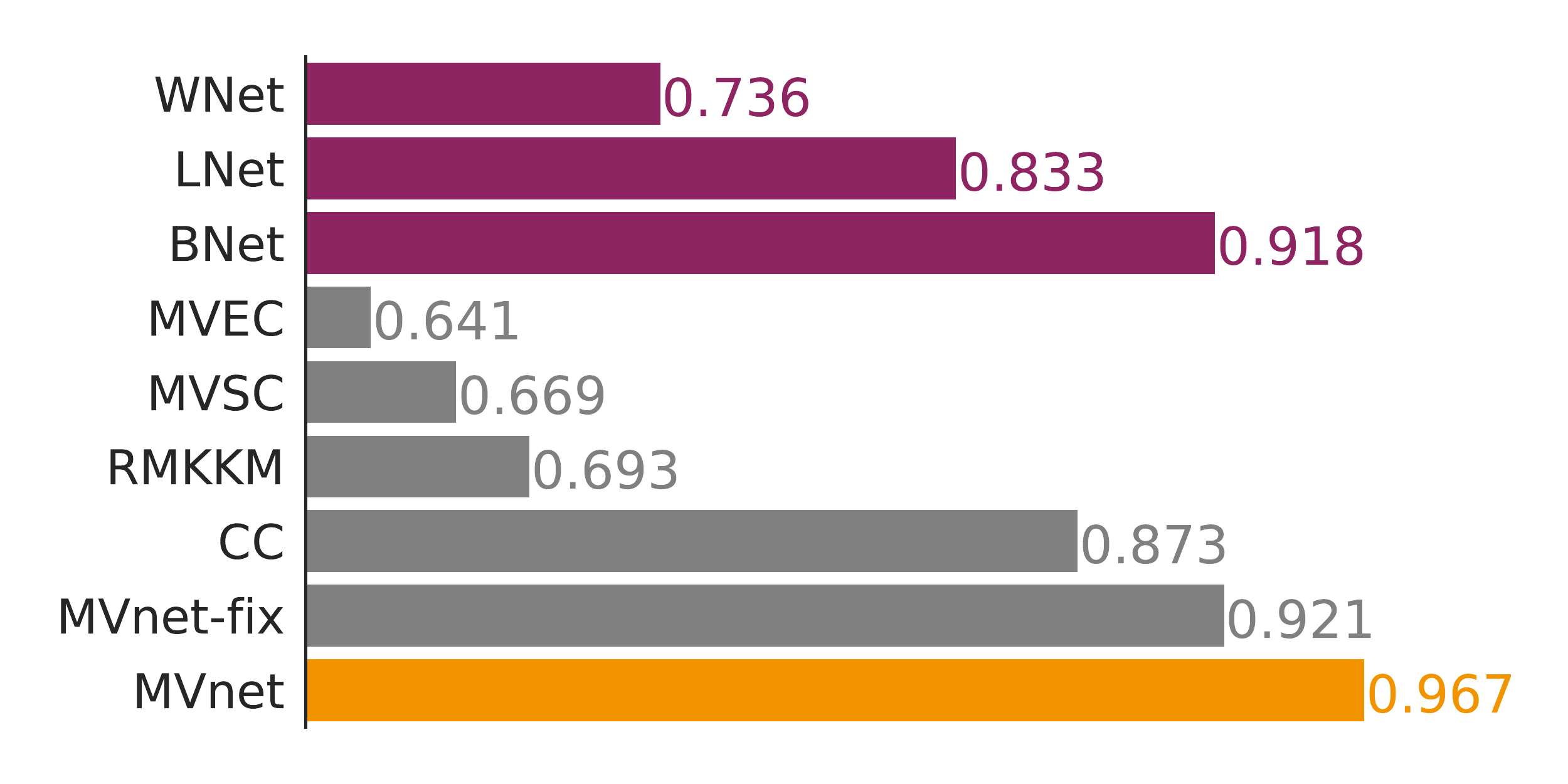}
\vspace{-25pt}
\caption{UMist}\label{fig:mvtc_allResults_umist}
\end{subfigure}

\begin{subfigure}[b]{0.45\textwidth}
\includegraphics[width=\textwidth]{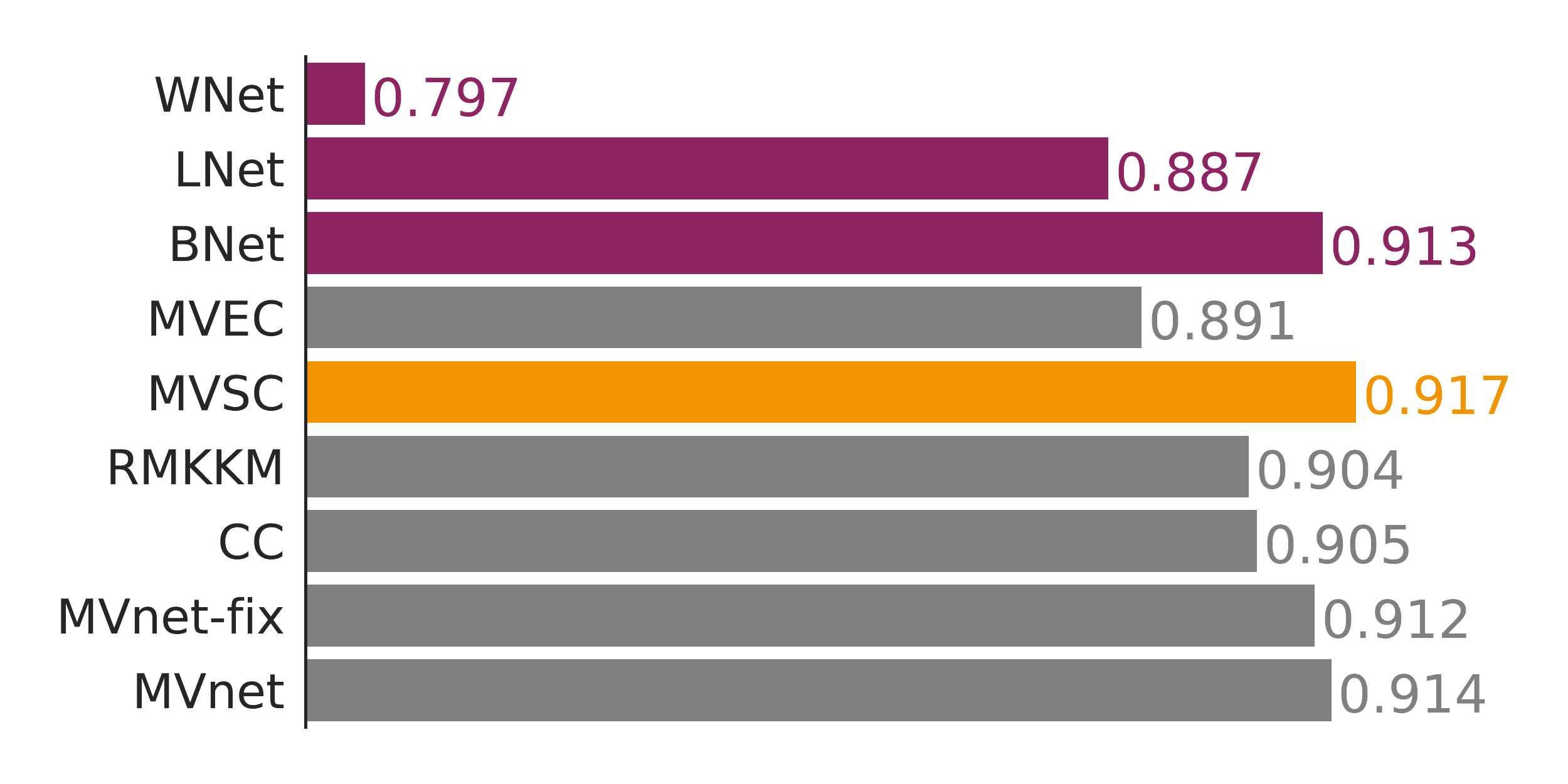}
\vspace{-25pt}
\caption{FEI}\label{fig:mvtc_allResults_fei}
\end{subfigure}

\caption{MIX score values for the different MVC methods applied to face datasets. \textit{Purple: single extractor,  gray: multi-view methods, orange: best}.}
\label{fig:mvtc_allResults_face}
\end{figure}

First of all, an algorithm is good if it produces cluster assignments with high NMI, purity and accuracy. To simplify the analysis of the results, we introduce a mixed clustering evaluation metric to analyze jointly NMI, PUR and ACC results:
\begin{equation}
    \text{MIX} =  \frac{\text{NMI} + \text{PUR} + \text{ACC}}{3}.
\end{equation}
Figures~\ref{fig:mvtc_allResults_object}, \ref{fig:mvtc_allResults_scene}, \ref{fig:mvtc_allResults_FG} and \ref{fig:mvtc_allResults_face} represent the MIX scores for the different methods and datasets. The results referenced as ``N/A'' could not be computed because the 16GB of RAM in the machine used were not sufficient to run some of the algorithms on large datasets.

To evaluate the interest of the proposed multi-view generation approach, all multi-view methods (DMVC, MVEC, MVSC, RMKKM) need to be compared with each feature extractor taken independently. Comparing results with each of the ten networks is both cumbersome and difficult to analyze. Instead, we prefer to report results for the Best network (BNet) and the Worst one (WNet). BNet (respectively WNet) represents the network which demonstrates the best (respectively worst) results on the precise dataset where it appears. In practice, BNet is impossible to choose in advance because on a true unsupervised dataset, external evaluation metrics (NMI, PUR, ACC, etc.) cannot be computed. The only possible strategy for selecting a feature extractor $f_z^{j^*}$ among $F_z = \{f_z^j, j \in \{1, ..., M\}\}$ is one we call the Leading Network (LNet) strategy. Given a clustering problem $X$ and a clustering algorithm $A$, the LNet strategy consists in using a set of $P$ supervised datasets $\{(X_1, y_1^*), ..., (X_P, y_P^*)\}$ and choose $f_z^{j^*}$ such that
\begin{equation}
j^* = \argmax\limits_{j \in \{1, ..., M\}} \left(\frac{1}{P} \sum\limits_{p = 1}^P \text{MIX}_{0.5}(A(f_z^j(X_p)), y_p^*)\right).
\end{equation}
In other words, the leading network is the one which presents the best clustering results on average across the $P$ supervised datasets, with respect to algorithm $A$. Using the online optimization formalism \cite{oco}, different feature extractors can be considered as experts and the obtained MIX scores as rewards from previous trials. Then, the LNet strategy simply becomes a Follow-the-Leader strategy. In our case, we use 9 datasets to evaluate the proposed image clustering methods. Hence, for each dataset, $f_z^{j^*}$ is computed by applying the LNet strategy on the $P = 8$ other datasets. The results obtained by $f_z^{j^*}$ on each dataset are reported under the name LNet. 

The condensed version of the results defined above can be found in Figures~\ref{fig:mvtc_allResults_object} to \ref{fig:mvtc_allResults_face}. We now propose to analyze these experimental results to explain three key findings.

\subsection{Results interpretations}
\label{sec:exp_mvc_res_interpretation}

\subsubsection{IC can benefit from the use of several CNN feature extractors}

When using multiple pretrained CNNs instead of one, the ideal scenario is when the MV approach outperforms every independent network, i.e. it outperforms BNet. When this occurs, we can conclude that the different feature extractors contain complementary information which should be leveraged when possible (see Figure~\ref{fig:intuitionEnsembles_union} and Section~\ref{sec:ic2mvc_intuition}). In Figures~\ref{fig:mvtc_allResults_object} to \ref{fig:mvtc_allResults_face}, we can see that for all datasets except CIFAR10, the best of all methods is a MV method.

We remind that, in practice, it is impossible to predict which feature extractor will be the BNet. When facing an unsupervised dataset, without additional knowledge, the only way to obtain the BNet results is to choose a CNN at random and to get lucky. The risk of using random selection is to fall in the worst case scenario (WNet). This risk can be measured by the margin separating the MV methods from WNet. Likewise, the potential benefit of random selection is measured by the difference between MV and BNet results. In the results, we see that random selection is not worth considering because the risk is much higher than the potential benefit (even for CIFAR10), which in most datasets tried is even negative.

The second possible benefit of leveraging our MV generation approach is to improve results from LNet. To the best of our knowledge, the LNet strategy introduced above is the only feature extractor selection method which is better than random. Figures~\ref{fig:mvtc_allResults_object} to \ref{fig:mvtc_allResults_face} show that MVnet is above LNet for all datasets except for flowers after retraining. We also point out that for all 9 datasets, there is at least one ensemble method that outperforms LNet. One possible way to improve the LNet strategy would be to increase the number of trials, i.e. the number of dataset on which LNet is computed. However, we doubt that the results would vary much with a larger $P$. These experimental results suggest that when facing an unsupervised dataset, using multiple feature extractors should be preferred over selecting a single one.

\subsubsection{MVC can be improved by adopting an end-to-end approach}

We first note that methods implemented with JULE outperform agglomerative clustering for most of the datasets. The three exceptions are the two scene recognition datasets (Archi and MIT) and Birds. One possible reason for the failure of JULE in these cases might be that the hyperparameters are not appropriate. Indeed, on the one hand, in the original paper, authors give hyperparameters recommendations for natural recognition and face recognition datasets. For these two tasks, we note that the results with JULE are very good. On the other hand, for other IC problems, such as scene and fine-grained recognition, different parameters may work better. The scalability of JULE to different kind of IC datasets would be worth investigating. We also believe that other deep clustering methods might work better on these datasets. For example, the results reported in \cite{alternating_opt_clust} appear to be good for unsupervised scene recognition tasks. It might be worth trying to adapt their method to multiview data and thus improve results on Archi and MIT.

To evaluate the interest of using MVnet independently from other considerations, such as the problems involved by JULE on certain datasets, we now only look at the 5 bottom bars of each diagram. The first thing we note is that the MVnet architecture is better suited for multiview data than the CC approach. We also underline that in most cases, CC presents limited interest compared to LNet. These results suggests that, for a new IC dataset, data extracted from multiple CNN feature extractors should be preprocessed independently before being considered jointly. We now compare the MVnet approach against standard multi-view approaches (MVEC, MVSC, RMKKM). Overall, MVnet seems to perform better and in cases where it does not, results are similar. We also note that MVnet results are much more consistent across the different datasets. For example, on COIL100, all three standard method present results below WNet, which is never the case for MVnet. Finally, fine-tuning MVnet$_{\text{fix}}$ end-to-end seem to be a good idea. Indeed, except for Flowers and CIFAR10, it always seem to perform either better or similarly.

\subsubsection{Combining multi-view generation from multiple architectures and DMVC produces state-of-the-art results at IC}

We conclude this results interpretation section by stating that, to the best of our knowledge, the results reported in this paper for VOC2007, COIL100, Flowers, UMist and FEI are the new state-of-the-art for classifying these datasets without labels.

\subsection{Learned representations}\label{sec:mvtc_exp_rep}

The quality of a deep clustering algorithm can also be assessed by studying the new feature representation it generates.

\subsubsection{Evaluation with K-means}
\begin{figure}
\centering

\includegraphics[width=0.48\textwidth]{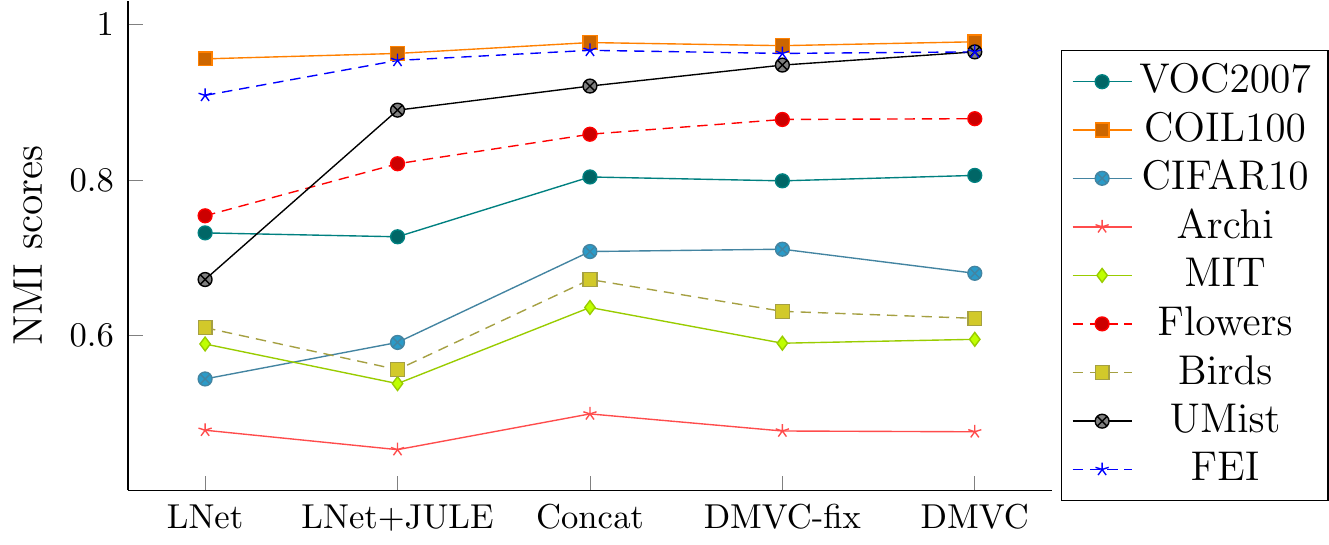}

\caption{NMI scores for K-means applied to feature representations from different stages of the DMVC pipeline.}
\label{fig:representationKmeans}
\end{figure}

The features extracted with MVnet are first evaluated by reclustering them using K-means. K-means is a simple clustering algorithm which performs best on representations presenting compact clusters, which are distant from each others. We choose the LNet to represent the fixed CNN feature representation methods. Results are reported in Figure~\ref{fig:representationKmeans} and show that for most datasets, better features are generated as we progress in the training of MVnet. For Birds, MIT and Archi, the results remain similar, which is likely to come from the fact that JULE does not perform well on these datasets.

\subsubsection{Visualization}
\begin{figure}
    \centering
    
    \begin{subfigure}[b]{0.23\textwidth}
        \centering \includegraphics[width=\textwidth]{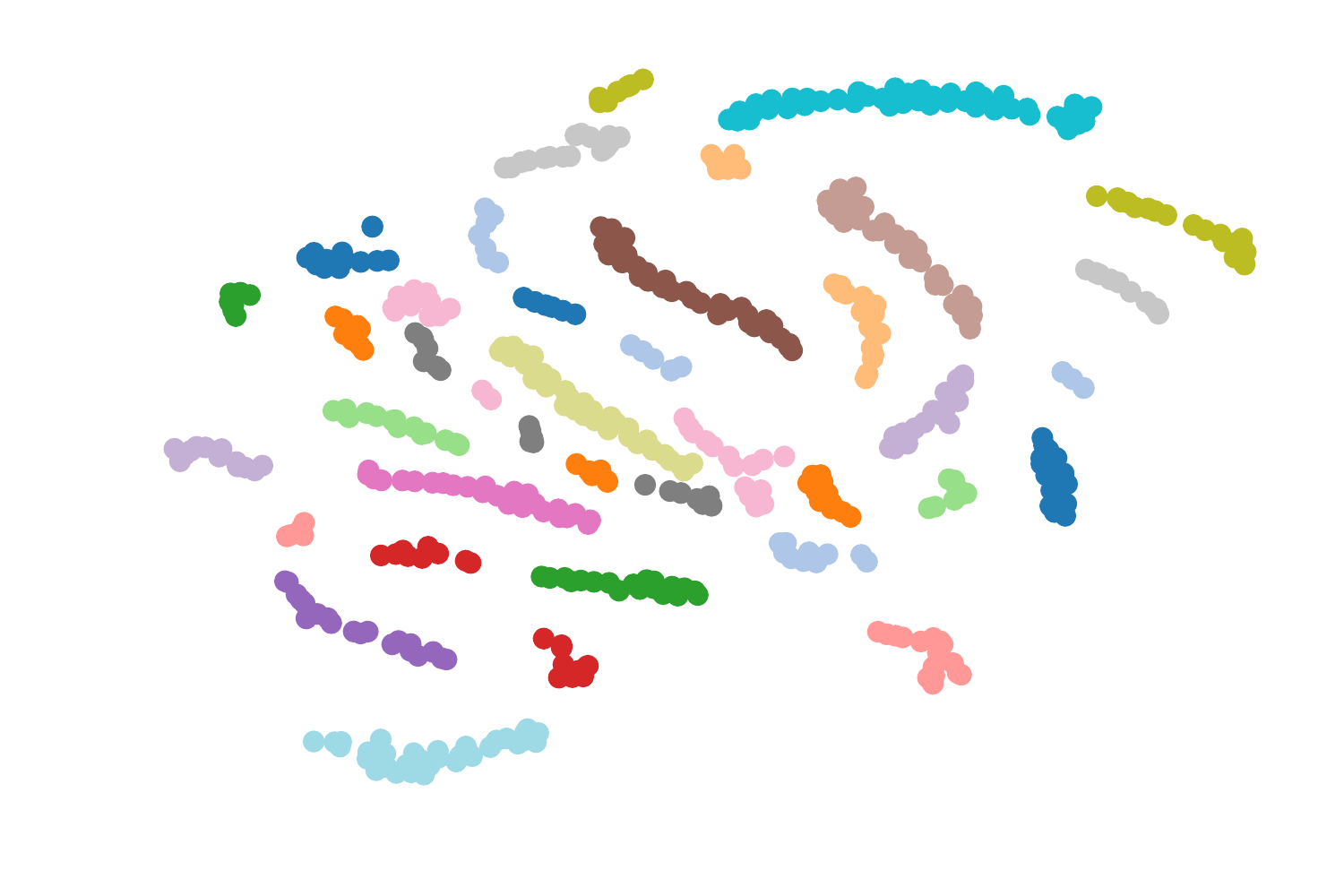}
        \caption{Densenet169 features}
        \label{fig:mvtc_representationTsne_dense}
    \end{subfigure}
    ~
    \begin{subfigure}[b]{0.23\textwidth}
        \centering \includegraphics[width=\textwidth]{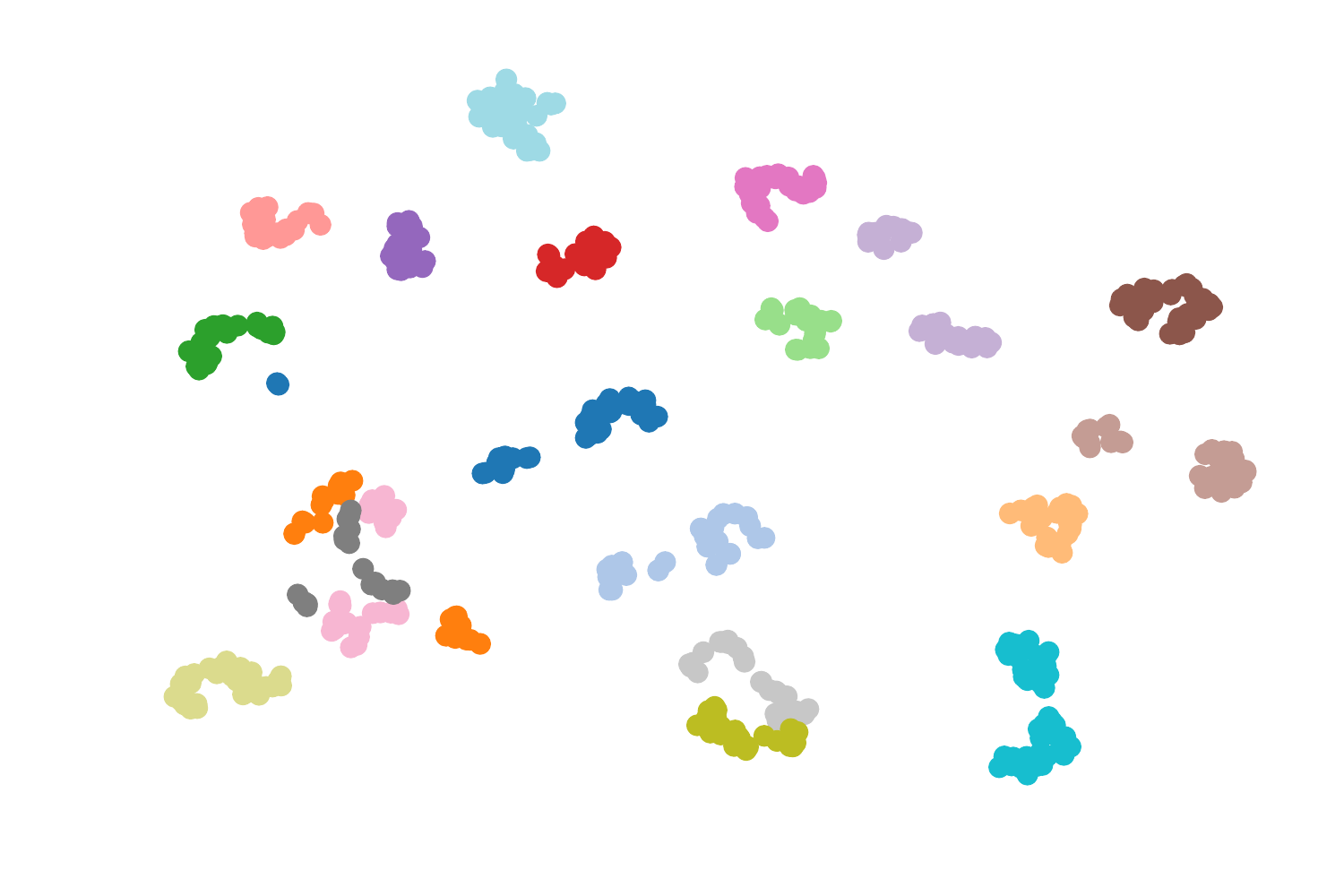}
        \caption{Densenet169 + JULE}
        \label{fig:mvtc_representationTsne_denseJule}
    \end{subfigure}
    
    \begin{subfigure}[b]{0.23\textwidth}
        \centering \includegraphics[width=\textwidth]{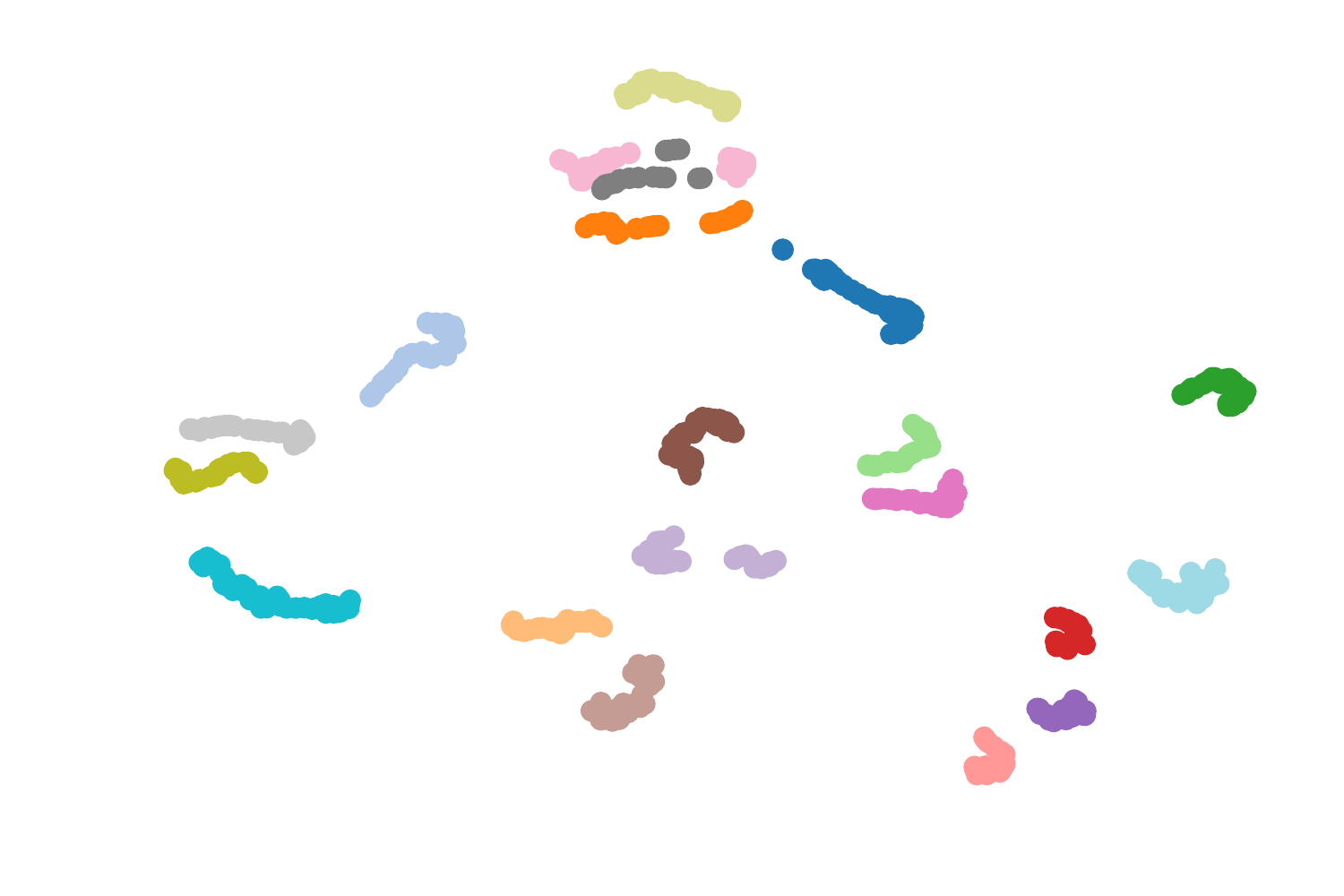}
        \caption{Concat}
        \label{fig:mvtc_representationTsne_concat}
    \end{subfigure}
    ~
    \begin{subfigure}[b]{0.23\textwidth}
        \centering \includegraphics[width=\textwidth]{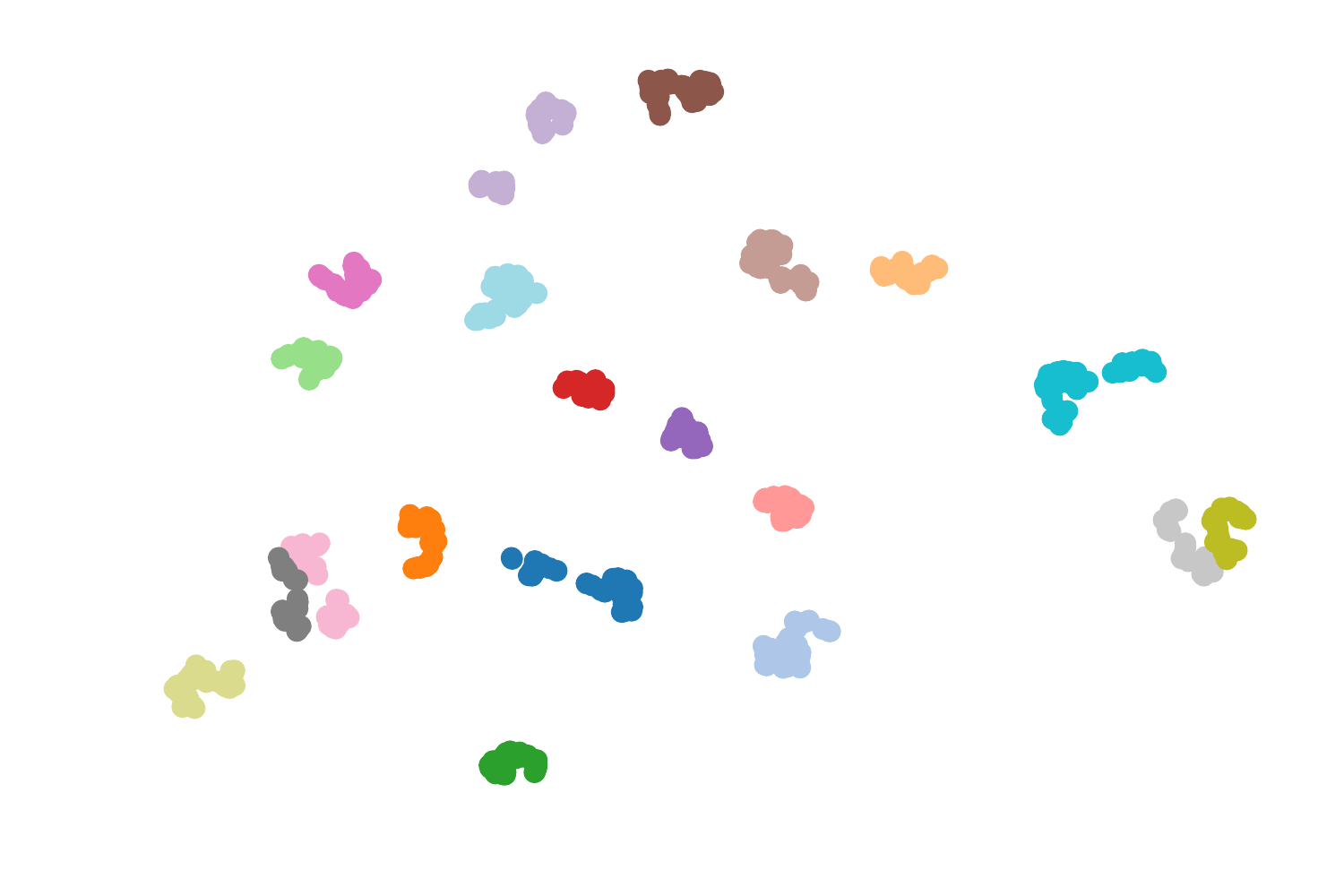}
        \caption{MVnet$_{\text{fix}}$}
        \label{fig:representationTsne_mvnetFix}
    \end{subfigure}
    
    \begin{subfigure}[b]{0.23\textwidth}
        \centering \includegraphics[width=\textwidth]{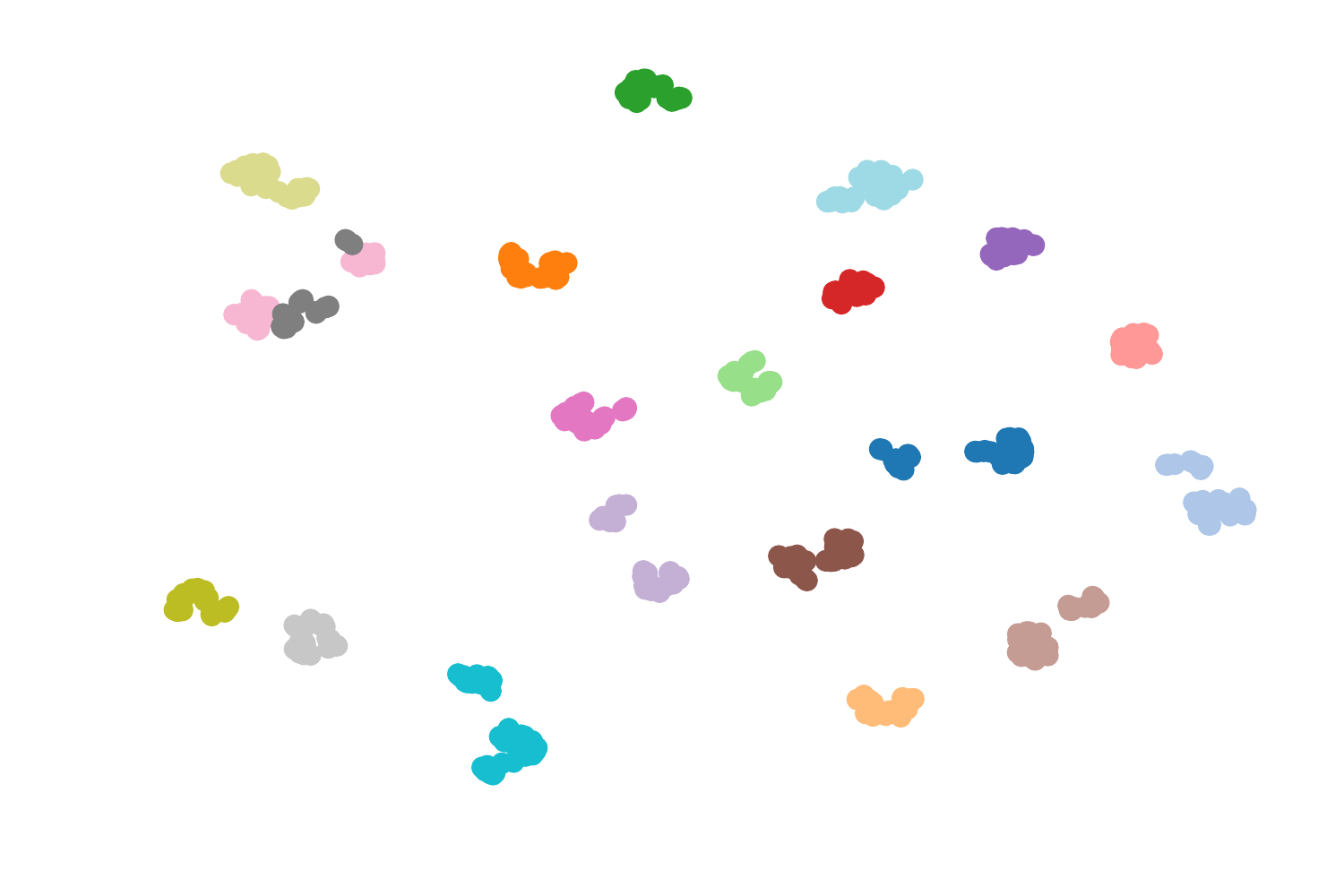}
        \caption{MVnet}
        \label{fig:representationTsne_mvnet}
    \end{subfigure}
    
    \caption[Visualization of the features generated by DMVC]{2d t-SNE visualization of the features extracted from the UMist dataset at different stages of the DMVC framework.\vspace{-4mm}}
    \label{fig:representationTsne}
\end{figure}

In Figure~\ref{fig:representationTsne}, we also propose to visualize the evolution of the 2d t-SNE representation of features at different stages of the MVnet training for the UMist dataset. It also shows that this way of training MVnet produces representations that generate more compact clusters, which are distant from each others.

\section{Conclusion and future works}
\label{sec:conclu}

\subsection{Conclusion}

The current state-of-the-art methods for complex datasets are using Convolutional Neural Networks pretrained on Imagenet to extract features from large natural images. However, the choice of the architecture and layer for feature extraction is often arbitrary. In this paper, we first conducted extended experiments on 8 standard computer vision datasets from four different IC subtasks to investigate the behavior of these features. Our first key finding is that for all architectures and tasks, the last layer before softmax seems to produce the most discriminative features for clustering. Our experiments also demonstrate that the selected deep feature extractor has a major impact on the results, however, they do not give any insight about how to select it. 

The importance of architecture selection, together with the absence of method to solve this problem, motivates the introduction of a new two-step approach to solve the image clustering problem. First, we generate multiple representations of each image using pretrained CNN feature extractors, and reformulate the problem as a multi-view clustering problem. Second, we define a multi-input neural network architecture, MVnet, which is used to solve MVC in an end-to-end manner. In theory, any deep clustering framework can be adapted to train an MVnet on unsupervised data. In practice we propose to implement this approach within the JULE framework and demonstrate state-of-the-art results for image clustering on several natural images datasets.

\subsection{Future research directions}

Our experimental results illustrate that different CNNs, pretrained on the same task, often contain different and complementary information about a target dataset. Differences may arise from a number of sources including the architecture (number of layers, layer shape, presence of skip connections, etc.), the regularization method, or the loss functions used for training. Investigating which parameters influence knowledge transfer to unsupervised tasks is an interesting axis of research for future work and such knowledge may help to design better CNN architectures.

We also note that pretrained CNNs are used as feature extractors for many applications, not just clustering. Using multiple pretrained CNNs to define a multi-view learning problem may be appealing for other tasks where complementary information present in pretrained feature extractors can improve performance.

Although JULE~\cite{jule} appears to be a good algorithm to solve MVC, we acknowledged that it fails for some datasets. Hence, it would be interesting to try different deep clustering methods or to adapt the parameters of JULE for different IC tasks.

Recently, self-supervised learning techniques~\cite{selfsupervised} have been used as an alternative to supervised training on Imagenet. They enable to learn good generic visual features on large unlabeled datasets, which can then be used to extract features to solve other tasks. In future work, it would be interesting to compare feature extractors pretrained on Imagenet against feature extractors trained with self-supervision. Combining them using our approach might also lead to even better results.

Finally, as the number of available pretrained CNNs will keep increasing and many researchers have limited resources, the case where the number of available feature extractors is much larger than the number of available GPUs needs to be considered. Hence, investigating an optimal strategy for architectures selection and resources allocation to maximize the available information is a promising research direction.

\section*{Acknowledgments}
This work was carried out under a Fulbright Haut-de-France Scholarship as well as a scholarship from the Fondation Arts et M\'etiers (convention No. 8130-01). This work is also supported by the European Union's 2020 research and innovation program under grant agreement No.688807, project ColRobot (collaborative robotics for assembly and kitting in smart manufacturing).


\newpage

\onecolumn
\appendix
\section{Complete result tables from Section~\ref{sec:benchmark_results} benchmark study}\label{annex:full_results_svtc}

\begin{table}[!ht]
\caption{Benchmark results on natural object recognition datasets.} 
\label{tab:svtc_res_natural}

\centering

\begin{adjustbox}{center}
\begin{tabular}{cc"cc|cc"cc|cc}

\multicolumn{2}{c"}{} & \multicolumn{4}{c"}{VOC2007} & \multicolumn{4}{c}{COIL100} \tabularnewline
\multicolumn{2}{c"}{} & \multicolumn{2}{c|}{KM} & \multicolumn{2}{c"}{Agg} & \multicolumn{2}{c|}{KM} & \multicolumn{2}{c}{Agg} \tabularnewline
\multicolumn{2}{c"}{} & nmi & \textit{time} & nmi & \textit{time} & nmi & \textit{time} & nmi & \textit{time} \tabularnewline

\Xhline{4\arrayrulewidth}

\multirow{3}{*}{VGG16}
& L1 & 0.516 & \textit{82.5} & 0.525 & \textit{96.8} & 0.895 & \textit{457} & 0.928 & \textit{476} \tabularnewline
& L2 & 0.578 & \textit{18.3} & 0.608 & \textit{16} & 0.943 & \textit{72} & 0.940 & \textit{71} \tabularnewline
& L3 & 0.673 & \textit{16.4} & 0.651 & \textit{16.1} & 0.945 & \textit{77} & 0.956 & \textit{70} \tabularnewline

\Xhline{4\arrayrulewidth}

\multirow{3}{*}{VGG19}
& L1 & 0.541 & \textit{83} & 0.537 & \textit{98} & 0.889 & \textit{440} & 0.922 & \textit{476} \tabularnewline
& L2 & 0.625 & \textit{17.6} & 0.618 & \textit{16.1} & 0.937 & \textit{67} & 0.949 & \textit{71} \tabularnewline
& L3 & 0.661 & \textit{18.5} & 0.650 & \textit{16.8} & 0.939 & \textit{62} & 0.948 & \textit{70} \tabularnewline

\Xhline{4\arrayrulewidth}

\multirow{3}{*}{Inception}
& L1 & 0.216 & \textit{{825}} & 0.298 & \textit{{856}} & 0.799 & \textit{{4,340}} & 0.846 & \textit{{4,455}} \tabularnewline
& L2 & 0.423 & \textit{{565}} & 0.542 & \textit{{531}} & 0.913 & \textit{{2,852}} & 0.939 & \textit{{2,517}} \tabularnewline
& L3 & 0.603 & \textit{{8.6}} & 0.692 & \textit{{8.6}} & 0.928 & \textit{{37}} & 0.953 & \textit{{35}} \tabularnewline

\Xhline{4\arrayrulewidth}

\multirow{3}{*}{Xception}
& L1 & 0.291 & \textit{{356}} & 0.410 & \textit{{403}} & 0.832 & \textit{{1,965}} & 0.875 & \textit{{1,960}} \tabularnewline
& L2 & 0.538 & \textit{{901}} & 0.625 & \textit{{822}} & 0.920 & \textit{{3,754}} & 0.942 & \textit{{3,909}} \tabularnewline
& L3 & 0.687 & \textit{{8.7}} & 0.719 & \textit{{8.1}} & 0.938 & \textit{{32}} & 0.955 & \textit{{35}} \tabularnewline

\Xhline{4\arrayrulewidth}

\multirow{3}{*}{Resnet50}
& L1 & 0.279 & \textit{{758}} & 0.315 & \textit{{783}} & 0.850 & \textit{{4,475}} & 0.888 & \textit{{3,876}} \tabularnewline
& L2 & 0.413 & \textit{{99}} & 0.494 & \textit{{97}} & 0.884 & \textit{{489}} & 0.927 & \textit{{474}} \tabularnewline
& L3 & 0.680 & \textit{{8.4}} & 0.656 & \textit{{8}} & 0.957 & \textit{{35}} & 0.967 & \textit{{35}} \tabularnewline

\Xhline{4\arrayrulewidth}

\multicolumn{2}{c"}{BoF} & 0.083 & \textit{{0.5}} & 0.072 & \textit{{0.4}} & - & -  \tabularnewline

\end{tabular}
\end{adjustbox}
\end{table}

\begin{table}[!ht]
\caption{Benchmark results on Scene recognition datasets.} 
\label{tab:svtc_res_scene}

\centering

\begin{adjustbox}{center}
\begin{tabular}{cc"cc|cc"cc|cc}

\multicolumn{2}{c"}{} & \multicolumn{4}{c"}{Archi} & \multicolumn{4}{c}{MIT} \tabularnewline
\multicolumn{2}{c"}{} & \multicolumn{2}{c|}{KM} & \multicolumn{2}{c"}{Agg} & \multicolumn{2}{c|}{KM} & \multicolumn{2}{c}{Agg} \tabularnewline
\multicolumn{2}{c"}{} & nmi & \textit{{time}} & nmi & \textit{{time}} & nmi & \textit{{time}} & nmi & \textit{{time}} \tabularnewline

\Xhline{4\arrayrulewidth}

\multirow{3}{*}{VGG16}
& L1 & 0.425 & \textit{{218}} & 0.420 & \textit{{210}} & 0.383 & \textit{{1,826}} & 0.410 & \textit{{2,259}} \tabularnewline
& L2 & 0.420 & \textit{{35}} & 0.431 & \textit{{31.2}} & 0.481 & \textit{{285}} & 0.471 & \textit{{328}} \tabularnewline
& L3 & 0.430 & \textit{{39}} & 0.414 & \textit{{31.4}} & 0.515 & \textit{{283}} & 0.492 & \textit{{334}} \tabularnewline

\Xhline{4\arrayrulewidth}

\multirow{3}{*}{VGG19}
& L1 & 0.416 & \textit{{219}} & 0.431 & \textit{{211}} & 0.388 & \textit{{1,749}} & 0.411 & \textit{{2,235}} \tabularnewline
& L2 & 0.415 & \textit{{37}} & 0.426 & \textit{{31.4}} & 0.484 & \textit{{288}} & 0.480 & \textit{{331}} \tabularnewline
& L3 & 0.408 & \textit{{34.1}} & 0.398 & \textit{{31.2}} & 0.510 & \textit{{289}} & 0.491 & \textit{{331}} \tabularnewline

\Xhline{4\arrayrulewidth}

\multirow{3}{*}{Inception}
& L1 & 0.184 & \textit{{1,747}} & 0.178 & \textit{{1,909}} & 0.209 & \textit{{170.4}} & 0.378 & \textit{{36,824}} \tabularnewline
& L2 & 0.412 & \textit{{1,142}} & 0.401 & \textit{{1,119}} & 0.495 & \textit{{10,005}} & 0.498 & \textit{{12,243}} \tabularnewline
& L3 & 0.420 & \textit{{16.1}} & 0.421 & \textit{{15.5}} & 0.570 & \textit{{138}} & 0.561 & \textit{{164}} \tabularnewline

\Xhline{4\arrayrulewidth}

\multirow{3}{*}{Xception}
& L1 & 0.165 & \textit{{595}} & 0.174 & \textit{{859}} & 0.430 & \textit{{7,307}} & 0.457 & \textit{{9,177}} \tabularnewline
& L2 & 0.435 & \textit{{1,636}} & 0.435 & \textit{{1,756}} & 0.500 & \textit{{174.8}} & 0.548 & \textit{{29,548}} \tabularnewline
& L3 & 0.442 & \textit{{15.5}} & 0.433 & \textit{{15.6}} & 0.597 & \textit{{121}} & 0.587 & \textit{{164}} \tabularnewline

\Xhline{4\arrayrulewidth}

\multirow{3}{*}{Resnet50}
& L1 & 0.248 & \textit{{1,827}} & 0.270 & \textit{{1,674}} & 0.193 & \textit{{165.7}} & 0.393 & \textit{{23949}} \tabularnewline
& L2 & 0.377 & \textit{{238}} & 0.389 & \textit{{210}} & 0.367 & \textit{{1,684}} & 0.395 & \textit{{2,252}} \tabularnewline
& L3 & 0.455 & \textit{{17}} & 0.447 & \textit{{22}} & 0.539 & \textit{{133}} & 0.529 & \textit{{164}} \tabularnewline

\Xhline{4\arrayrulewidth}

\multicolumn{2}{c"}{BoF} & 0.100 & \textit{{0.9}} & 0.102 & \textit{{1.1}} & - & - \tabularnewline

\end{tabular}
\end{adjustbox}
\end{table}

\begin{table}[!ht]
\caption{Benchmark results on fine-grained recognition datasets.} 
\label{tab:svtc_res_FG}

\centering

\begin{adjustbox}{center}
\begin{tabular}{cc"cc|cc"cc|cc}

\multicolumn{2}{c"}{} & \multicolumn{4}{c"}{Flowers} & \multicolumn{4}{c}{Birds} \tabularnewline
\multicolumn{2}{c"}{} & \multicolumn{2}{c|}{KM} & \multicolumn{2}{c"}{Agg} & \multicolumn{2}{c|}{KM} & \multicolumn{2}{c}{Agg} \tabularnewline
\multicolumn{2}{c"}{} & nmi & \textit{{time}} & nmi & \textit{{time}} & nmi & \textit{{time}} & nmi & \textit{{time}} \tabularnewline

\Xhline{4\arrayrulewidth}

\multirow{3}{*}{VGG16}
& L1 & 0.588 & \textit{{36.5}} & 0.616 & \textit{{17.1}} & 0.375 & \textit{{2,309}} & 0.420 & \textit{{1,288}} \tabularnewline
& L2 & 0.657 & \textit{{5.5}} & 0.661 & \textit{{2.5}} & 0.529 & \textit{{399}} & 0.542 & \textit{{190}} \tabularnewline
& L3 & 0.688 & \textit{{5.7}} & 0.644 & \textit{{2.7}} & 0.557 & \textit{{371}} & 0.557 & \textit{{189}} \tabularnewline

\Xhline{4\arrayrulewidth}

\multirow{3}{*}{VGG19}
& L1 & 0.567 & \textit{{35}} & 0.614 & \textit{{17.1}} & 0.390 & \textit{{2,369}} & 0.428 & \textit{{1,282}} \tabularnewline
& L2 & 0.652 & \textit{{5.5}} & 0.676 & \textit{{2.5}} & 0.534 & \textit{{377}} & 0.544 & \textit{{189}} \tabularnewline
& L3 & 0.652 & \textit{{5.3}} & 0.646 & \textit{{2.5}} & 0.562 & \textit{{383}} & 0.557 & \textit{{187}} \tabularnewline

\Xhline{4\arrayrulewidth}

\multirow{3}{*}{Inception}
& L1 & 0.241 & \textit{{364}} & 0.329 & \textit{{155}} & 0.143 & \textit{{145.7}} & 0.306 & \textit{{17,966}} \tabularnewline
& L2 & 0.560 & \textit{{219}} & 0.604 & \textit{{90.2}} & 0.464 & \textit{{12,483}} & 0.484 & \textit{{6,904}} \tabularnewline
& L3 & 0.650 & \textit{{2.6}} & 0.677 & \textit{{1.1}} & 0.558 & \textit{{164}} & 0.569 & \textit{{95}} \tabularnewline

\Xhline{4\arrayrulewidth}

\multirow{3}{*}{Xception}
& L1 & 0.425 & \textit{{139}} & 0.495 & \textit{{69.7}} & 0.341 & \textit{{8,919}} & 0.366 & \textit{{5,194}} \tabularnewline
& L2 & 0.633 & \textit{{298}} & 0.661 & \textit{{140}} & 0.481 & \textit{{189.8}} & 0.548 & \textit{{11,159}} \tabularnewline
& L3 & 0.674 & \textit{{2.9}} & 0.670 & \textit{{1.2}} & 0.633 & \textit{{159}} & 0.644 & \textit{{94}} \tabularnewline

\Xhline{4\arrayrulewidth}

\multirow{3}{*}{Resnet50}
& L1 & 0.543 & \textit{{298}} & 0.589 & \textit{{137}} & 0.146 & \textit{{202.3}} & 0.331 & \textit{{10,994}} \tabularnewline
& L2 & 0.510 & \textit{{37.6}} & 0.607 & \textit{{17.2}} & 0.375 & \textit{{2,630}} & 0.421 & \textit{{1,273}} \tabularnewline
& L3 & 0.684 & \textit{{2.7}} & 0.708 & \textit{{2.3}} & 0.521 & \textit{{197}} & 0.529 & \textit{{94}} \tabularnewline

\Xhline{4\arrayrulewidth}

\multicolumn{2}{c"}{BoF} & 0.177 & \textit{{0.331}} & 0.179 & \textit{{0.103}} & - & - \tabularnewline

\end{tabular}
\end{adjustbox}
\end{table}

\begin{table}[!ht]
\caption{Benchmark results on face recognition datasets.} 
\label{tab:svtc_res_face}

\centering

\begin{adjustbox}{center}
\begin{tabular}{cc"cc|cc"cc|cc}

\multicolumn{2}{c"}{} & \multicolumn{4}{c"}{Umist} & \multicolumn{4}{c}{FEI} \tabularnewline
\multicolumn{2}{c"}{} & \multicolumn{2}{c|}{KM} & \multicolumn{2}{c"}{Agg} & \multicolumn{2}{c|}{KM} & \multicolumn{2}{c}{Agg} \tabularnewline
\multicolumn{2}{c"}{} & nmi & \textit{{time}} & nmi & \textit{{time}} & nmi & \textit{{time}} & nmi & \textit{{time}} \tabularnewline

\Xhline{4\arrayrulewidth}

\multirow{3}{*}{VGG16}
& L1 & 0.575 & \textit{{11.9}} & 0.699 & \textit{{3.2}} & 0.875 & \textit{{304}} & 0.909 & \textit{{72}} \tabularnewline
& L2 & 0.669 & \textit{{1.7}} & 0.717 & \textit{{0.5}} & 0.899 & \textit{{52}} & 0.900 & \textit{{11}} \tabularnewline
& L3 & 0.669 & \textit{{1.5}} & 0.689 & \textit{{1.5}} & 0.887 & \textit{{45}} & 0.897 & \textit{{11}} \tabularnewline

\Xhline{4\arrayrulewidth}

\multirow{3}{*}{VGG19}
& L1 & 0.623 & \textit{{12.7}} & 0.717 & \textit{{3.1}} & 0.888 & \textit{{311}} & 0.923 & \textit{{93}} \tabularnewline
& L2 & 0.719 & \textit{{1.5}} & 0.766 & \textit{{0.4}} & 0.905 & \textit{{47}} & 0.918 & \textit{{15}} \tabularnewline
& L3 & 0.700 & \textit{{1.5}} & 0.740 & \textit{{0.4}} & 0.908 & \textit{{45}} & 0.915 & \textit{{13}} \tabularnewline

\Xhline{4\arrayrulewidth}

\multirow{3}{*}{Inception}
& L1 & 0.588 & \textit{{104}} & 0.641 & \textit{{28.1}} & 0.881 & \textit{{3,366}} & 0.920 & \textit{{663}} \tabularnewline
& L2 & 0.661 & \textit{{60.9}} & 0.694 & \textit{{16.5}} & 0.923 & \textit{{1,843}} & 0.939 & \textit{{383}} \tabularnewline
& L3 & 0.698 & \textit{{0.8}} & 0.760 & \textit{{0.2}} & 0.921 & \textit{{21}} & 0.941 & \textit{{5.3}} \tabularnewline

\Xhline{4\arrayrulewidth}

\multirow{3}{*}{Xception}
& L1 & 0.589 & \textit{{47.2}} & 0.642 & \textit{{12.9}} & 0.789 & \textit{{1,393}} & 0.831 & \textit{{296}} \tabularnewline
& L2 & 0.688 & \textit{{89.9}} & 0.745 & \textit{{25.5}} & 0.930 & \textit{{2,712}} & 0.934 & \textit{{599}} \tabularnewline
& L3 & 0.683 & \textit{{0.8}} & 0.731 & \textit{{0.2}} & 0.913 & \textit{{20}} & 0.928 & \textit{{5.2}} \tabularnewline

\Xhline{4\arrayrulewidth}

\multirow{3}{*}{Resnet50}
& L1 & 0.533 & \textit{{87.9}} & 0.597 & \textit{{24.9}} & 0.843 & \textit{{3,137}} & 0.892 & \textit{{579}} \tabularnewline
& L2 & 0.555 & \textit{{11.3}} & 0.623 & \textit{{3.1}} & 0.903 & \textit{{324}} & 0.931 & \textit{{72}} \tabularnewline
& L3 & 0.658 & \textit{{770}} & 0.717 & \textit{{0.2}} & 0.910 & \textit{{25}} & 0.919 & \textit{{5.3}} \tabularnewline

\Xhline{4\arrayrulewidth}

\multicolumn{2}{c"}{BoF} & 0.542 & \textit{{0.306}} & 0.638 & \textit{{0.03}} & - & - \tabularnewline

\end{tabular}
\end{adjustbox}
\end{table}

\clearpage

\section{Complete result tables from Section~\ref{sec:exp_mvc_res}}\label{annex:full_results_mvtc}

\begin{table}[!ht]
\caption{Results on natural object datasets.
(\textit{N/A}: \textit{Insufficient memory to compute})} 
\label{tab:mvtc_res_natural}

\centering

\scalebox{1}{
\begin{adjustbox}{center}
\begin{tabular}{c|ccc|ccc|ccc}

& \multicolumn{3}{c|}{VOC2007} & \multicolumn{3}{c}{COIL100} & \multicolumn{3}{c}{CIFAR10}
\tabularnewline
& NMI & PUR & ACC & NMI & PUR & ACC & NMI & PUR & ACC \tabularnewline
\hline

VGG16 & 0.695 & 0.761 & 0.743 & 0.994 & 0.970 & 0.961 & 0.456 & 0.474 & 0.474  \tabularnewline

VGG19 & 0.676 & 0.734 & 0.710 & 0.994 & 0.978 & 0.971 & 0.477 & 0.424 & 0.424
\tabularnewline

Inception & 0.764 & 0.819 & 0.791 & 0.988 & 0.948 & 0.938 & 0.608 & 0.599 & 0.599
\tabularnewline
 
Xception & 0.763 & 0.785 & 0.775 & 0.988 & 0.949 & 0.936 & 0.539 & 0.444 & 0.443 \tabularnewline
  
Resnet50 & 0.719 & 0.762 & 0.736 & 0.966 & 0.980 & 0.975 & 0.554 & 0.487 & 0.486 \tabularnewline
   
Densenet121 & 0.748 & 0.778 & 0.736 & 0.991 & 0.964 & 0.957 & 0.563 & 0.535 & 0.535 \tabularnewline
 
Densenet169 & 0.741 & 0.791 & 0.776 & 0.993 & 0.970 & 0.962 & 0.557 & 0.47 & 0.47 \tabularnewline

Densenet201 & 0.769 & 0.800 & 0.788 & 0.993 & 0.970 & 0.957 & 0.62 & 0.59 & 0.59 \tabularnewline

InceptionResnet & 0.763 & 0.809 & 0.800 & 0.982 & 0.928 & 0.916 & 0.646 & 0.608 & 0.608 \tabularnewline

Nasnet & 0.752 & 0.811 & 0.777 & 0.979 & 0.906 & 0.882 & 0.264 & 0.336 & 0.332
\tabularnewline

MVEC & 0.794 & 0.816 & 0.787 & 0.967 & 0.874 & 0.845 & \textit{N/A} & \textit{N/A} & \textit{N/A} \tabularnewline

MVEC$_{\text{jule}}$ & 0.820 & 0.813 & 0.812 & 0.996 & 0.980 & 0.975 & 0.62 & 0.47 & 0.469 \tabularnewline

MVSC & 0.740 & 0.811 & 0.662 & 0.955 & 0.868 & 0.839 & \textit{N/A} & \textit{N/A} & \textit{N/A} \tabularnewline

MVSC$_{\text{jule}}$ & 0.772 & 0.824 & 0.649 & 0.984 & 0.952 & 0.943 & \textit{N/A} & \textit{N/A} & \textit{N/A} \tabularnewline

RMKKM & 0.759 & 0.826 & 0.679 & 0.954 & 0.858 & 0.827 & \textit{N/A} & \textit{N/A} & \textit{N/A} \tabularnewline

RMKKM$_{\text{jule}}$ & 0.773 & 0.831 & 0.717 & 0.976 & 0.91 & 0.881 & \textit{N/A} & \textit{N/A} & \textit{N/A} \tabularnewline
 
CC & 0.745 & 0.806 & 0.785 & 0.996 & 0.980 & 0.971 & 0.574 & 0.569 & 0.569 \tabularnewline

MVnet$_{fix}$ & 0.816 & 0.826 & 0.803 & 0.995 & 0.980 & 0.971 & 0.654 & 0.558 & 0.558 \tabularnewline

MVnet & 0.827 & 0.860 & 0.814 & 0.996 & 0.980 & 0.973 & 0.6 & 0.46 & 0.46

\end{tabular}
\end{adjustbox}}
\end{table}

\begin{table}[!ht]
\centering

\caption{Results on scene datasets.
(\textit{N/A}: \textit{Insufficient memory to compute})} 
\label{tab:mvtc_res_scene}

\scalebox{1}{
\begin{adjustbox}{center}
\begin{tabular}{c|ccc|ccc}

& \multicolumn{3}{c|}{Archi} & \multicolumn{3}{c}{MIT}
\tabularnewline
& NMI & PUR & ACC & NMI & PUR & ACC \tabularnewline
\hline

VGG16 & 0.386 & 0.331 & 0.314 &  0.471 & 0.385 & 0.373 \tabularnewline

VGG19 & 0.397 & 0.349 & 0.341 & 0.483 & 0.422 & 0.401 \tabularnewline

Inception & 0.422 & 0.343 & 0.317 & 0.517 & 0.43 & 0.412 \tabularnewline
 
Xception & 0.445 & 0.402 & 0.372 & 0.572 & 0.483 & 0.469 \tabularnewline
  
Resnet50 & 0.412 & 0.333 & 0.323 & 0.494 & 0.404 & 0.372 \tabularnewline
   
Densenet121 & 0.451 & 0.376 & 0.346 & 0.502 & 0.406 & 0.387 \tabularnewline
 
Densenet169 & 0.432 & 0.367 & 0.33 & 0.53 & 0.462 & 0.433 \tabularnewline

Densenet201 & 0.461 & 0.431 & 0.412 & 0.463 & 0.346 & 0.332 \tabularnewline

InceptionResnet & 0.408 & 0.346 & 0.324 & 0.543 & 0.473 & 0.46 \tabularnewline

Nasnet & 0.428 & 0.362 & 0.334 & 0.613 & 0.524 & 0.509 \tabularnewline

MVEC & 0.505 & 0.479 & 0.447 & 0.644 & 0.564 & 0.537 \tabularnewline

MVEC$_{\text{jule}}$ & 0.471 & 0.362 & 0.35 & 0.6 & 0.454 & 0.446 \tabularnewline

MVSC & 0.478 & 0.48 & 0.421 & \textit{N/A} & \textit{N/A} & \textit{N/A} \tabularnewline

MVSC$_{\text{jule}}$ & 0.495 & 0.476 & 0.406 & \textit{N/A} & \textit{N/A} & \textit{N/A} \tabularnewline

RMKKM & 0.454 & 0.461 & 0.378 & \textit{N/A} & \textit{N/A} & \textit{N/A} \tabularnewline

RMKKM$_{\text{jule}}$ & 0.479 & 0.489 & 0.422 & \textit{N/A} & \textit{N/A} & \textit{N/A} \tabularnewline
 
CC & 0.449 & 0.378 & 0.367 & 0.53 & 0.43 & 0.408 \tabularnewline

MVnet$_{fix}$ & 0.478 & 0.432 & 0.403 & 0.588 & 0.491 & 0.476 \tabularnewline

MVnet & 0.487 & 0.449 & 0.433 & 0.591 & 0.495 & 0.482

\end{tabular}
\end{adjustbox}}
\end{table}

\begin{table}[!ht]
\centering

\caption{Results on fine-grained datasets.
(\textit{N/A}: \textit{Insufficient memory to compute})}
\label{tab:mvtc_res_FG}

\scalebox{1}{
\begin{adjustbox}{center}
\begin{tabular}{c|ccc|ccc}

& \multicolumn{3}{c|}{Flowers} & \multicolumn{3}{c}{Birds}
\tabularnewline
& NMI & PUR & ACC & NMI & PUR & ACC \tabularnewline
\hline

VGG16 & 0.703 & 0.658 & 0.654 & 0.493 & 0.158 & 0.148 \tabularnewline

VGG19 & 0.677 & 0.642 & 0.629 & 0.513 & 0.192 & 0.18 \tabularnewline

Inception & 0.758 & 0.718 & 0.717 & 0.538 & 0.199 & 0.187 \tabularnewline
 
Xception & 0.705 & 0.675 & 0.662 & 0.604 & 0.278 & 0.263 \tabularnewline
  
Resnet50 & 0.718 & 0.69 & 0.676 & 0.44 & 0.129 & 0.12 \tabularnewline
   
Densenet121 & 0.832 & 0.801 & 0.798 & 0.552 & 0.193 & 0.182 \tabularnewline
 
Densenet169 & 0.831 & 0.815 & 0.813 & 0.523 & 0.184 & 0.171 \tabularnewline

Densenet201 & 0.81 & 0.795 & 0.793 & 0.53 & 0.184 & 0.166 \tabularnewline

InceptionResnet & 0.594 & 0.535 & 0.527 & 0.462 & 0.146 & 0.129 \tabularnewline

Nasnet & 0.659 & 0.584 & 0.579 & 0.495 & 0.18 & 0.167 \tabularnewline

MVEC & 0.798 & 0.731 & 0.727 & 0.664 & 0.343 & 0.336 \tabularnewline

MVEC$_{\text{jule}}$ & 0.849 & 0.758 & 0.756 & 0.573 & 0.178 & 0.168 \tabularnewline

MVSC & 0.765 & 0.752 & 0.734 & \textit{N/A} & \textit{N/A} & \textit{N/A}\tabularnewline

MVSC$_{\text{jule}}$ & 0.87 & 0.876 & 0.869 & \textit{N/A} & \textit{N/A} & \textit{N/A}\tabularnewline

RMKKM & 0.77 & 0.764 & 0.755 & \textit{N/A} & \textit{N/A} & \textit{N/A}\tabularnewline

RMKKM$_{\text{jule}}$ & 0.868 & 0.874 & 0.868 & \textit{N/A} & \textit{N/A} & \textit{N/A}\tabularnewline
 
CC & 0.761 & 0.712 & 0.707 & 0.563 & 0.229 & 0.216 \tabularnewline

MVnet$_{fix}$ & 0.879 & 0.86 & 0.859 & 0.617 & 0.271 & 0.258 \tabularnewline

MVnet & 0.833 & 0.766 & 0.757 & 0.612 & 0.273 & 0.258

\end{tabular}
\end{adjustbox}}
\end{table}

\begin{table}[!ht]
\caption{Results on face datasets.} 
\label{tab:mvtc_res_face}

\centering

\scalebox{1}{
\begin{adjustbox}{center}
\begin{tabular}{c|ccc|ccc}

& \multicolumn{3}{c|}{UMist} & \multicolumn{3}{c}{FEI}
\tabularnewline
& NMI & PUR & ACC & NMI & PUR & ACC \tabularnewline
\hline

VGG16 & 0.891 & 0.802 & 0.776 & 0.924 & 0.775 & 0.748 \tabularnewline

VGG19 & 0.879 & 0.798 & 0.717 & 0.934 & 0.798 & 0.775 \tabularnewline

Inception & 0.841 & 0.722 & 0.666 & 0.953 & 0.854 & 0.837 \tabularnewline
 
Xception & 0.834 & 0.732 & 0.642 & 0.953 & 0.85 & 0.837 \tabularnewline
  
Resnet50 & 0.959 & 0.908 & 0.887 & 0.953 & 0.869 & 0.86 \tabularnewline
   
Densenet121 & 0.906 & 0.845 & 0.798 & 0.959 & 0.895 & 0.886 \tabularnewline
 
Densenet169 & 0.933 & 0.88 & 0.838 & 0.959 & 0.879 & 0.867 \tabularnewline

Densenet201 & 0.904 & 0.819 & 0.776 & 0.957 & 0.861 & 0.843 \tabularnewline

InceptionResnet & 0.889 & 0.803 & 0.75 & 0.913 & 0.748 & 0.73 \tabularnewline

Nasnet & 0.886 & 0.809 & 0.762 & 0.946 & 0.819 & 0.803 \tabularnewline

MVEC & 0.763 & 0.61 & 0.551 & 0.949 & 0.865 & 0.858 \tabularnewline

MVEC$_{\text{jule}}$ & 0.942 & 0.877 & 0.854 & 0.966 & 0.903 & 0.897 \tabularnewline

MVSC & 0.753 & 0.654 & 0.6 & 0.955 & 0.901 & 0.895 \tabularnewline

MVSC$_{\text{jule}}$ & 0.912 & 0.847 & 0.793 & 0.972 & 0.944 & 0.944 \tabularnewline

RMKKM & 0.773 & 0.675 & 0.631 & 0.959 & 0.882 & 0.872 \tabularnewline

RMKKM$_{\text{jule}}$ & 0.923 & 0.875 & 0.83 & 0.963 & 0.902 & 0.887 \tabularnewline
 
CC & 0.936 & 0.87 & 0.812 & 0.96 & 0.882 & 0.872 \tabularnewline

MVnet$_{fix}$ & 0.966 & 0.922 & 0.875 & 0.962 & 0.891 & 0.882 \tabularnewline

MVnet & 0.984 & 0.967 & 0.95 & 0.963 & 0.893 & 0.885

\end{tabular}
\end{adjustbox}}
\end{table}

\end{document}